\documentclass[lettersize,journal]{IEEEtran}

\usepackage{microtype}
\usepackage{url}
\usepackage{xcolor}
\usepackage{combelow}
\newcommand{\chrupala}{Chrupa\l a}

\usepackage{tikz}
\usepackage{mathtools}
\usepackage{fontawesome}
\usetikzlibrary{backgrounds}
\usetikzlibrary{positioning}
\usetikzlibrary{fit}

\usepackage{amsmath}
\usepackage{amssymb}
\usepackage{namedtensor}
\usepackage{stmaryrd}
\DeclareMathOperator*{\mean}{mean}

\DeclareMathOperator*{\argmax}{argmax}
\ndef{\temp}{T}
\ndef{\emb}{K}
\ndef{\vocab}{W}

\newcommand{\ab}{\mathbf{a}}  
\newcommand{\ib}{\mathbf{i}}  
\newcommand{\tb}{\mathbf{t}}  
\newcommand{\hb}{\mathbf{h}}  
\newcommand{\yb}{\mathbf{y}}  
\newcommand{\Hb}{\mathbf{H}}  
\newcommand{\yv}{y^\mathrm{vis}} 
\newcommand{\yt}{y^\mathrm{bow}} 
\newcommand{\ybv}{\mathbf{y}^{\mathrm{vis}}}
\newcommand{\dnna}{\mathrm{DNN}_a} 
\newcommand{\dnni}{\mathrm{DNN}_i} 
\newcommand{\enc}{\mathrm{Enc}} 
\newcommand{\pool}{\mathrm{Pool}} %
\newcommand{\clf}{\mathrm{Clf}}

\usepackage{xspace}
\newcommand{\etal}{{\it et al.}\xspace}

\usepackage{cite}  
\usepackage{booktabs}
\usepackage{tabularx}
\usepackage{multirow}
\usepackage{siunitx}
\newcolumntype{C}{>{\centering\arraybackslash}X}
\newcolumntype{L}{>{\raggedright\arraybackslash}X}
\newcolumntype{R}{>{\raggedleft\arraybackslash}X}

\newcommand{\ii}[1]{{\footnotesize \textcolor{gray}{#1}}}

\definecolor{hermancolor}{HTML}{FF6600}
\definecolor{dancolor}{HTML}{9A00FF}
\definecolor{kayodecolor}{HTML}{00BFFF}

\usepackage[prependcaption,textsize=scriptsize]{todonotes}
\begin{document}
    
\title{Keyword localisation in untranscribed speech \\ using visually grounded speech models}
    
\author{
    Kayode Olaleye, Dan Onea\cb{t}\u{a}, and Herman Kamper
    \thanks{%
        This work is supported in part by the National Research Foundation of South Africa (grant no. 120409), a Google Africa PhD Scholarship,
        and a grant of the Romanian Ministry of Education and Research (CNCS-UEFISCDI, project number PN-III-P1-1.1-PD-2019-0918, within PNCDI III).%
    }
}

\markboth{In submission}
{Olaleye {\it et al.}}
    
    
\maketitle
    
\begin{abstract}
    Keyword localisation is the task of finding where in a speech utterance a given query keyword occurs.
    We investigate to what extent keyword localisation is possible using a visually grounded speech (VGS) model.
    VGS models are trained on unlabelled images paired with spoken captions. These models are therefore
    self-supervised---trained without any
    explicit textual label or location information.
    To obtain training targets, we first tag training images 
    with soft text labels using a pretrained visual classifier with a fixed vocabulary.
    This enables a VGS model to predict the presence of a written keyword in an utterance, but not its location.
    We consider four ways to equip VGS models with localisations capabilities.
    Two of these---a saliency approach and input masking---can be applied to an arbitrary prediction model after training, while the other two---attention and a score aggregation approach---are incorporated directly into the structure of the model.
    Masked-based localisation gives some of the best reported localisation scores from a VGS model, with an accuracy of 57\% when the system knows that a keyword occurs in an utterance and need to predict its location.
    In a setting where localisation is performed after detection, an $F_1$ of 25\% is achieved, and in a setting where a keyword spotting ranking pass is first performed, a localisation $P@10$ of 32\% is obtained.
    While these scores are modest compared to the idealised setting with unordered bag-of-word-supervision (from transcriptions), these VGS
    models
    do not receive any textual or location supervision.
    Further analyses show that these models are limited by the first detection or ranking pass.
    Moreover, individual keyword localisation performance is correlated with the tagging performance from the visual classifier.
    We also show qualitatively how and where semantic mistakes occur, e.g. that the model locates \textit{surfer} when queried with \textit{ocean}.
\end{abstract}
\begin{IEEEkeywords}
    Visually grounded speech models, keyword localisation, keyword spotting, self-supervised learning.
\end{IEEEkeywords}

\section{Introduction}
\label{sec:introduction}

\IEEEPARstart{A}{utomatic}
speech recognition (ASR) enables human-computer interaction and improves accessibility by transcribing audio media.
However, accurate ASR systems are available in only a fraction of the world's 
languages because such systems require a vast amount of labelled data. 
As a result, there has been growing interest in speech processing systems that, instead of using exact transcriptions, can learn from weakly labelled data~\cite{duong2016, palaz2016, settle2017, weiss2017}.
One paradigm is the visually grounded speech (VGS) model that learns from images and their spoken captions~\cite{driesen2010, synnaeve2014b, harwath2015, harwath2016,  harwath2017, harwath2018a, harwath2018b, eloff2019, harwath2019a, harwath2019b}.
Since paired speech and images are signals available during early language acquisition, such co-occurring audio-visual inputs are ideal for modelling infants' language learning~\cite{bomba1983, pinker1994, eimas+quinn94, roy2003, boves2007, chrupala2016, okko2019}.
VGS models are also suitable for
teaching new words to robots~\cite{meng2013, nortje2020}.
Furthermore, 
images and utterances describing them could arguably be easier to collect than speech and transcriptions,
especially when developing systems for low-resource languages~\cite{de1998} or languages with no written form~\cite{scharenborg2018, lupke2010, bird2020, scharenborg2020, wang2021a}.

    VGS models 
    can be regarded as self-supervised: 
we have access to two views of the data (visual and auditory) without any human labels, and we learn transferable representations by maximising the agreement between the two modalities~\cite{tsai2021}.
One approach is to use a contrastive loss~\cite{harwath2016,chrupala2017,harwath2018b,havard2019a,havard2019b} 
that simultaneously enforces views for the same data point to be similar and views from different data points to be different.
    Here we follow another approach based on the work of Kamper~\textit{et al.}~\cite{kamper2017a, kamper2019b, pasad2019},
    who showed that we can build VGS models capable of detecting whether a written keyword occurs in an utterance without relying on any text annotations.
    The key ingredient 
    is the availability of a pretrained image tagger, which is used to supervise the speech model.
    More precisely, given an image, the visual tagger generates soft labels for a fixed vocabulary of visual categories,
    which the speech model tries to mimic on the corresponding spoken utterance.
    At test time, the speech model is then able to map an input utterance to a set of textual words that are likely to appear in the utterance.
    The numerous large-scale labelled image datasets~\cite{imagenet2009,xiao2010,lin2014,krishna2017,kuznetsova2020} and performant vision models~\cite{he2016,tan2019,dosovitskiy2021,tolstikhin2021} ensure the availability of the pretrained visual component.

    In this paper we go one step further and investigate to what extent we are able to localise where the detected keyword appears in the speech utterance. I.e., we consider the task of 
    \textit{keyword localisation} using VGS models.
    We underline that this is an incredibly challenging problem since neither textual labels nor any location information is provided at training time.
    
    One application of visually grounded keyword localisation is in the documentation of endangered languages~\cite{ferrand2020}.
    A linguist can employ such tools to rapidly locate the speech segments containing a query keyword in a collection of utterances;
    the only prerequisite is a collection of image-speech pairs from the target language to train the model.
    Moreover, the localisation techniques that we explore are also readily applicable to the case of bag-of-words annotations, where we have
    a partial labelled set of the words appearing in an utterance 
    but not their order, number of occurrences or location~\cite{palaz2016}.

    Since no location information is provided at training time, we have to make use of the intrinsic ability of the VGS models to localise \cite{oquab2015}.
    We investigate four localisation methods.
    The first
    involves a saliency-based method that highlights the input region that is the most important for a predicted output.
    In particular, we employ the popular explainability method Grad-CAM~\cite{selvaraju2017}, which was originally developed in the context of images, and adapt it to our task.
    The second approach involves masking the input signal at different locations and measuring the response score predicted by the trained model on the partial inputs;
    large variations in the output suggest the presence of a keyword.
    These first two approaches can be applied to any convolutional prediction model after training, irrespective of architecture.
    The third and fourth methods use the fact that for certain architectures the internal activations can be interpreted as localisation scores.
    Concretely, the third approach assumes that the final pooling layer is a simple transformation (e.g.\ average or max pooling),
    so that the output score can be regarded as an aggregation of local scores.
    These can be use to
    select the most likely temporal location for the query keyword.
    Finally, the fourth approach assumes that the network involves an attention layer that pools features over the temporal axis; we use the attention weights as localisation scores.

    As suggested, the above localisation approaches vary in the degree to which they are coupled with a network's architecture.
    For this reason we conduct a careful evaluation that takes into account the architecture of the speech network.
    We explore four variants differing in the choice of speech encoder (either by using max pooling layers or not) and final pooling layer (log-mean-exp, max pooling or attention).
    All the network variants are based on one-dimensional convolutional layers, as these have shown strong results on related audio tasks \cite{peddinti2015,dai2017,shon2018,snyder2018}.
    We compare the architectures in terms of both localisation and detection performance.
    Since detection is the first step in localisation in some settings, we also investigate the
    the impact of detection on subsequent
    localisation performance.

    This paper integrates and extends our previous work from \cite{olaleye2020,olaleye2021}.
    We specifically make the following new contributions:
    (1)~a unified and detailed description of the VGS framework, applied to both detection and localisation;
    (2)~a new localisation method, consisting of the masked-in and masked-out variants;
    (3)~a systematic and fair comparison of four different keyword localisation methods;
    (4)~more detailed experiments revealing the limits of the detection step for localisation, the impact of the architecture, and performance at the individual keyword level; and
    (5)~a qualitative analysis showcasing predicted locations and common failure cases.

\section{Related work}
\label{sec:related_Work}

The last few years have seen major developments in visually grounded speech (VGS) modelling  (Section~\ref{subsec:visually-grounded-speech-models}).
Much of this VGS work takes inspiration from earlier work on learning from images and text (Section~\ref{subsec:learning-from-images-and-text}).
Our work here also relates more broadly to studies in machine learning  using knowledge distillation (Section~\ref{subsec:cross-modal-distillation}).

\subsection{Visually grounded speech models}
\label{subsec:visually-grounded-speech-models}

\chrupala~\cite{chrupala2021} provides an exhaustive survey of datasets, architectures, downstream tasks, and evaluation techniques for VGS models over the last 20 years.
Here we only briefly highlight previous work that is most relevant to our study.

Synnaeve {\it et al.}~\cite{synnaeve2014b}
proposed one of the earliest multimodal networks consisting of a convolutional image branch and a feedforward speech branch that maps speech and images into a common space using a cosine-distance loss.
Harwath {\it et al.}~\cite{harwath2016} scaled up this approach using a much larger dataset in a network with convolutions in both branches.
They also introduced a contrastive margin loss that pushes matching image-speech pairs closer to each other compared to mismatched image-speech pairs; variants of this loss has since become the standard in most VGS model.
Other network architectures have been explored: \chrupala~\etal~\cite{chrupala2017} used a gated recurrent neural network with attention for the speech branch, while recently Peng and Harwath~\cite{peng2021} proposed a Transformer-based model.
These types of VGS models have been analysed in a variety of ways~\cite{chrupala2016,drexler2017,harwath2019a,havard2019b,scholten2021} and they have been extended to non-English and multilingual variants~\cite{harwath2018b,kamper2018,havard2019a}.
Most of these studies use the VGS models for retrieving images using a spoken utterance and vice-versa.
Very recent studies have also considered speech synthesis from images~\cite{wang2021b} and image generation from spoken captions~\cite{wang2021a}.

While models that can define a joint image-speech speech (or models that can move between spaces) are useful in themselves, they do not provide a mapping of speech to textual labels.
This was addressed by Kamper \textit{et al.}~\cite{kamper2017a, kamper2019b},
which we use as the starting point of our VGS models in this work.
As mentioned in Section~\ref{sec:introduction}, these VGS models are able to detect whether a written keyword occurs in an utterance.
They do so by using a pretrained visual tagger that generates soft labels for the training images, which then serves as labels for a convolutional neural network that maps speech to soft unordered word targets.
Pasad {\it et al.}~\cite{pasad2019} showed that this approach is even complementary in settings where (limited) text labels are available.
However, although these studies have illustrated that keywords can be detected, the question remains of whether these VGS models can also locate query keywords---we address this here.

\subsection{Learning from images and text}
\label{subsec:learning-from-images-and-text}

Many of the VGS studies mentioned above were preceded by models that learn from images and text.
Early studies considered defining a joint image-text space~\cite{socher2010,weston2011}.
In image captioning the goal is to produce a natural language description of a visual scene~\cite{farhadi2010,kulkarni2011,young2014,karpathy2015,bernardi2016,sharma2018}.
Most relevant to us, studies have shown that it is possible to link specific keywords in a caption to specific image regions~\cite{liu2017, lu2018, Rohrbach2018, selvaraju2019}.
In some cases keywords are sparsely connected to image segments, e.g.\ systems only locating noun words~\cite{plummer2015}, while in other cases all the words in a sentence need to be linked to an image region~\cite{ponttuset2020}.
These approaches, however, require alignment information during training, specifying a relevant image region for words in a caption.
We focus on a setting where written captions are unavailable during training: only images and their spoken captions are available.
Moreover, we do not have alignment information.
Instead, we let our system search different segments of speech at test time for the most probable segment containing a specified query keyword.

\subsection{Cross-modal distillation}
\label{subsec:cross-modal-distillation}

By having a teacher network (the visual model) to supervise a student network (the audio model), our work can be regarded as a knowledge distillation technique \cite{hinton2015,ba2013}.
The original methods used large networks to train smaller ones, but new applications have been emerging; see \cite{gou2021} for a recent survey.
Compared to the initial papers, we share similarities in how knowledge is transferred (via the logits produced by the teacher) and by the fact that the teacher is fixed (offline), but we differ in that our method operates across modalities.
This idea falls under the category of cross-modal distillation, which has been previously applied to perform image classification on depth data \cite{gupta2016}, pose estimation on radio signals \cite{zhao2018}, scene detection on audio sound \cite{aytar2016}, emotion recognition on speech \cite{albanie2018}.
Similar to our work, all these cases leverage a visual network for supervision,
but the roles can be reversed as done by Owens \etal \cite{owens2016phd,owens2018}, which use sound information to supervise the visual network or,
more generally, as in \cite{alwassel2020}, which learn two networks (audio and visual) in tandem on the pseudo-labels obtained by clustering the features from the other modality.
The representations learnt by these methods were successfully transferred to tasks such as sound classification~\cite{aytar2016,alwassel2020}, video classification~\cite{alwassel2020}, and object or scene recognition~\cite{owens2018}.
In our case,
our downstream tasks are related to speech, specifically keyword detection and keyword localisation.

\section{Keyword localisation in a VGS model}
\label{sec:vgs_models}

    In this section we present the methodological contributions.
    We start with an overview of the setup, describing both training and testing of the VGS models.
	We then give specific details of the different
 proposed methods for keyword localisation.

    \subsection{Methodological overview}
    \label{subsec:methodological-overview}

    The main component of our work is a system that, given an audio utterance $\ab$ and a keyword $w$,
    estimates the probability of the keyword occurring in that utterance.
    The system is implemented in terms of an audio deep neural network $\dnna$ that outputs a vector of predictions $\dnna(\ab) \in \mathbb{R}^\vocab$, each element corresponding to a score for one of the $\vocab$ words in the system vocabulary.
   	This vector is then passed through the sigmoid function $\sigma$ to ensure that the predictions are in $[0, 1]$ and can be interpreted as probabilities:   
    \begin{equation}
        \hat \yb \triangleq p(w|\ab) = \sigma(\dnna(\ab)).
    \end{equation}
    To obtain the probability for a single keyword $w$, we select the entry from $\hat \yb \in [0, 1]^\vocab$ corresponding to the $w$-th keyword in the vector, i.e. $\hat{y}_w$.

\begin{figure}[t]
    \centering
    \includegraphics[width=0.99\columnwidth]{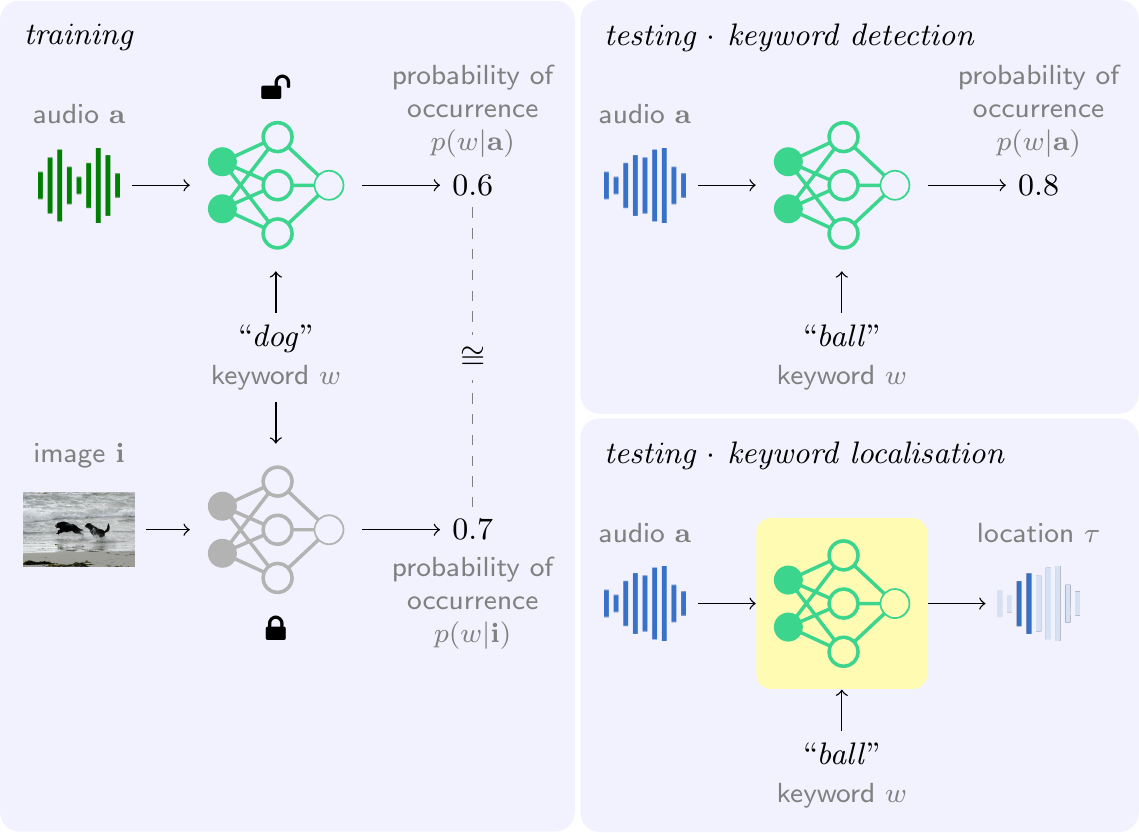}
    \caption{%
        Methodological overview.
        \textit{Left:} We train an audio neural network to predict whether a keyword $w$ occurs in an audio utterance $\ab$
        based on the supervision of a fixed pretrained image tagger ran on a corresponding image $\ib$.
        \textit{Top-right:}
        At test time, the network can be used
        to identify whether a certain keyword occurs in any given utterance (keyword detection).
        \textit{Bottom-right:} It can also be used
        to locate the keyword in the input (keyword localisation).
    }
    \label{fig:overview}
\end{figure}

     \textbf{Network structure.}
    We assume that the audio network $\dnna$ consists of three components that are sequenced---%
    an audio encoder $\enc$, a pooling layer $\pool$, and a classifier $\clf$:
    \begin{equation}
        \dnna = \clf \circ \pool \circ \enc.
    \end{equation}
    The audio encoder $\enc$ produces a sequence of $\emb$-dimensional embeddings $\left[\hb_1, \dots, \hb_\temp\right] = \Hb \in \mathbb{R}^{\emb \times \temp}$,
    which are then aggregated across the temporal dimension (of $\temp$ steps) by the pooling layer.
The result is finally passed through the classifier $\clf$ to obtain a $\vocab$-dimensional vector corresponding to the scores of the keywords.

    \textbf{Training.}
    To train the model, we use \textit{visual supervision}.
    Concretely, we assume that:
    (1) for each utterance in the training data, there is a corresponding matching image, and
    (2)~we have access to a pretrained image tagger to generate probability labels from images over the same vocabulary of $\vocab$ words.
    The target outputs 
    $\ybv \in [0, 1]^\vocab$ for the audio network will be the soft labels produced by the image tagger: an image deep neural network $\dnni$.
    For the image $\ib$ associated with an utterance $\ab$, we therefore obtain:
    \begin{equation}
        \ybv \triangleq p(w|\ib) = \sigma(\dnni(\ib)).
    \end{equation}
    We optimise the binary cross-entropy between the predictions of the audio network (taking $\ab$ as input) and the predictions from the image network (taking $\ib$ as input):

    \begin{equation}
        L = \yv_w \log \hat y_w + (1 - \yv_w) \log (1 - \hat y_w).
        \label{eq:cross_entropy}
    \end{equation}
    $L$ is averaged over the keywords and audio-image pairs,
    and it is optimised with respect to the weights of the audio network,
    while the weights of the image network are frozen.
    This training procedure is illustrated on the left in Figure~\ref{fig:overview}.

    The ground truth provided by the visual branch is weak in two senses:
    (1) there is no localisation information---the word can appear anywhere in the utterance---and
    (2) the labels can be possibly erroneous since the audio description can be different from the visual information present in the image. 
	To measure the impact of (2), i.e.\ the impact of using possibly incorrect labels from the visual tagger,
    we also experiment with a case where the ground truth is
    a
    bag-of-words (BoW).
    A BoW labelling indicates
	the presence or absence of words without giving their location, order, or number of occurrences.
    In this scenario we generate the target labels $\yt$ from the text transcriptions $\tb$ of the audio sample, i.e.\ $\yt_w = \llbracket w \in \tb \rrbracket$, 
    where $\llbracket - \rrbracket$ denotes the indicator function.
    For this setting we use the same binary cross-entropy loss as in Equation~\ref{eq:cross_entropy}, but with the 
    BoW
     labels.

    \textbf{Testing: Keyword detection.}
    At test time the audio neural network can be used in the same manner as it was trained:
    given an audio utterance and a keyword, pass them through the network to obtain the probability of the occurrence of the specified keyword. 
	Note that no visual information is required at test time.
    We can obtain a binary decision regarding the presence of the keyword $w$ by thresholding the probability score $\hat y_w$ with a value $\theta_w$,
    that is
    $\llbracket \hat y_w \ge \theta_w \rrbracket$.
    This is illustrated on the top-right of Figure~\ref{fig:overview}.

    \textbf{Testing: Keyword localisation.}
    The main goal of the paper involves the task of keyword localisation:
    given an audio utterance and a keyword, find where in the input the keyword occurs.
    The chief idea is to perform localisation with the same audio network which was trained for classification---
i.e.\ the network is trained without  keyword location labels.
    We propose four types of methods that can adapt a keyword detection model for localisation;
    each of them is described in the next subsection. 
    To compute
    location, each of these methods assign a score $\alpha_{w,t}$ for each keyword $w$ and location $t$ of the input audio sequence.
    The predicted location $\tau_w$ for the input keyword $w$ is the position of the highest score across time:
    \begin{equation}
        \tau_w = \argmax \left\{ \alpha_{w,1}, \dots, \alpha_{w,\temp} \right\}.
        \label{eq:location-max}
    \end{equation}
    This is illustrated on the bottom-right of Figure~\ref{fig:overview}.
    
    \subsection{Methods for keyword localisation}
    \label{subsec:methods-for-keyword-localisation}

    In this subsection we describe the proposed methods for keyword localisation.
    We assume we are given an audio network trained for the task of keyword detection (Section~\ref{subsec:methodological-overview}). We present four classes of methods that are able to augment the output of the network with localisation scores.
    
    \subsubsection{Grad-CAM}~
    \label{subsubsec:grad-cam}

	Grad-CAM~\cite{selvaraju2017} is a saliency-based method that can be applied to any convolutional neural network to determine which parts of the network input most contribute to a particular output decision.
	Intuitively, it works by determining how important each filter in a convolutional layer is to a particular output prediction (a specified keyword in our case).
	Localisation scores are then calculated for an input utterance by weighing the filter activations with their importance for the keyword under consideration.
	We use the scores computed after the last convolutional layer as localisation scores.

	More formally,
    Grad-CAM first determines how important each filter $k$ in the output of the last convolutional layer $\hb_t$ is to the word $w$:
    \begin{equation}
        \gamma_{k, w} = \frac{1}{\temp} \sum^{\temp}_{t = 1} \frac{\partial \hat{y}_w}{\partial h_{t, k}}.
    \end{equation}
    The output activation $\hb_t$ at time step $t$ is then weighted by its importance $\mathbf\gamma_w$ towards word $w$, giving the localisation 
    scores:
    \begin{equation}
        \alpha_{w, t} = \mathrm{ReLU} \left[ \sum_{k = 1}^{\emb} \gamma_{k, w} h_{t, k} \right].
    \end{equation}
	
	\begin{figure}[bt]
		\centering
		\includegraphics[width=0.99\linewidth]{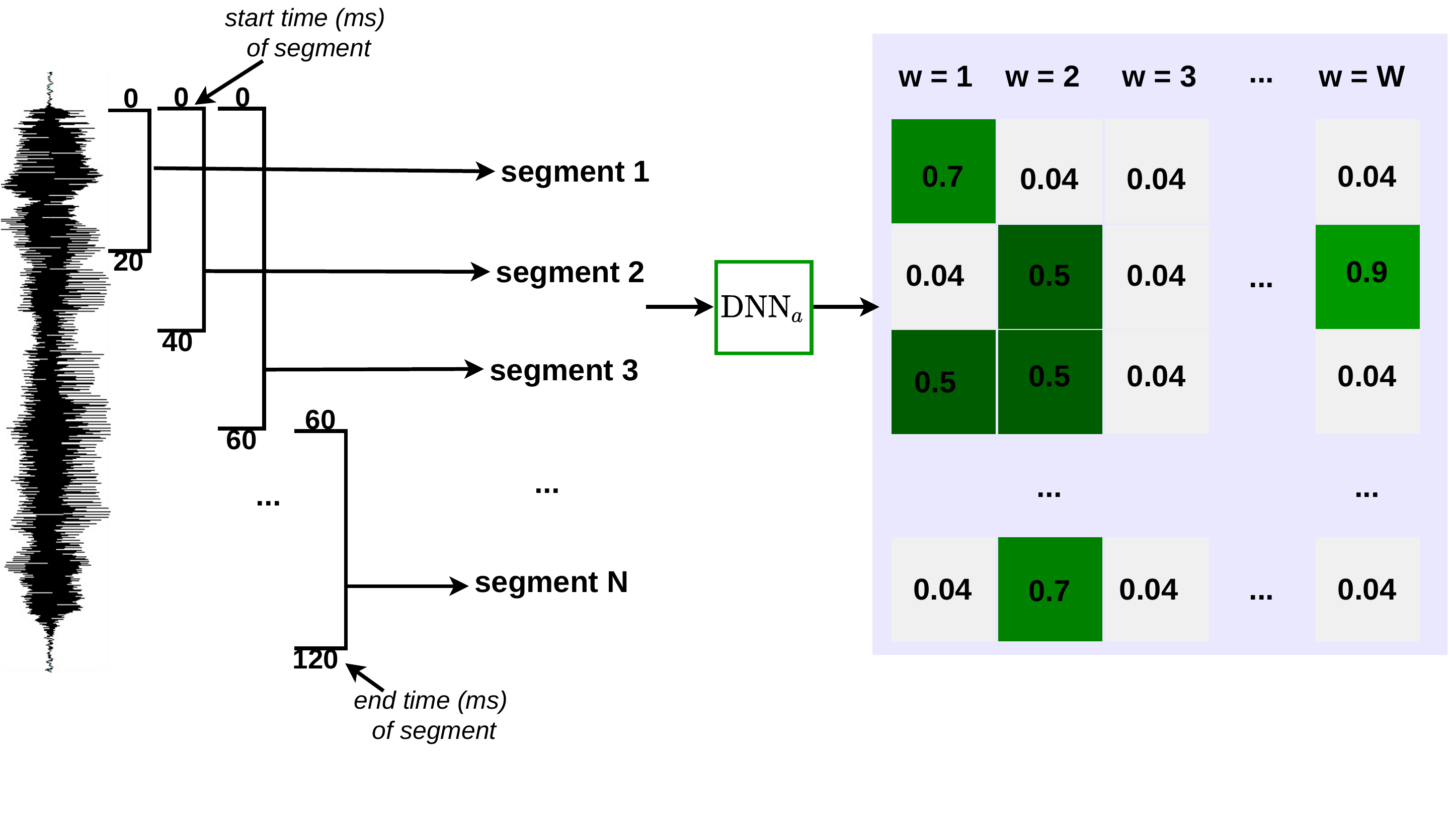}
		\caption{Our masked-in approach to keyword localisation in speech. An input utterance is split into overlapping segments of varying durations. Each segment is then fed into the network ($\dnna$) to determine the output detection probability of a particular word. The segment corresponding to the maximum output detection probability of the word is a possible location of the word. For instance, the most probable segment containing the word with index $w=1$ is segment 1, with $\alpha_{1, 1} = 0.7$.
		}
		\label{fig:masking_localisation_approach}
	\end{figure}

    \subsubsection{Input masking}~
    \label{subsubsec:input-masking}
    
    Rather than looking at intermediate activations within a network, an alternative is to modify a network's input at specific locations and then observe the impact on the output detection probabilities.
    For instance, in the \textit{masked-in} approach, we pass only a part of an input utterance through the model; if this increases the output probability of a particular word, that could be an indication that the word occurs in the segment.
    In the \textit{masked-out} approach, we instead pass an entire utterance through a network but mask out a part of the input; if this decreases the output probability of a particular word, that could be an indication that the word occurs in the masked-out segment.
    Both methods have the advantage that they can be applied to any type of detection model. 
    However, depending on how segments in an utterance are masked in or out, they 
    can be computationally expensive since for a single utterance we might need to make several passes through the model.
	
    More formally, the 
    \textit{masked-in} approach, as depicted in Figure~\ref{fig:masking_localisation_approach}, treats the keyword localisation task as a keyword detection task.
    Specifically, we split a test utterance into segments of minimum duration of 200 ms and maximum duration of 600 ms, with consecutive segments overlapping by 30 ms.
    Each segment is zero-padded
    back to the actual length of the utterance.
    We then feed each padded segment into the trained model to compute a detection score for a keyword.
    This process yields the following localisation scores:
    \begin{equation}
        \alpha_{w, t} = \sigma\left(\dnna(\ab \odot \mathbf{m}_t)_w\right),
    \end{equation}
    where $\mathbf{m}_t$ is a binary mask with ones to the entries corresponding to the $t$-th segment and zeros otherwise,
    and $\odot$ denotes element-wise multiplication.
	The segment for which the trained model produces the highest detection score for the keyword is treated as the location of the keyword.

    The \textit{masked-out} approach is the direct opposite, where we occlude (by replacing it with zeros) each segment from the utterance to compute the detection score.
    If the masked-out segment contains the keyword, we expect a detection score that is close to $0$.
    In order to select the location $\tau$ based on the maximum (Equation~\ref{eq:location-max}) we use the complement of the occurrence probabilities as localisation scores:
    \begin{equation}
        \alpha_{w, t} = 1 - \sigma\left(\dnna(\ab \odot (1 - \mathbf{m}_t))_w\right).
    \end{equation}

    \textbf{Relation to prior work.}
        The masked-in method operates similarly to the sliding window approach often used for object detection in images~\cite{viola2001}.
        An important distinction is that in our case the network that is repeatedly applied to each window is trained in a very weak setting with noisy supervision labels and without any location information.
    The masked-out approach is related to the work of Zeiler and Fergus~\cite{zeiler2014}; they 
investigated whether convolutional neural network-based vision models are really 
identifying the location of an object in an image or just finding
the surrounding context. 
We adapted their method of systematically perturbing different portions of the input image, but apply this approach to speech here.

In the speech domain,
Harwath and Glass~\cite{harwath2017} proposed a method for connecting word-like acoustic units of an utterance to semantically relevant image regions. They
did this
by extracting segments from the utterance and then used their  multimodal network to associate them with an appropriate subregion of the image. This is different from our setting because rather than matching speech segments to relevant image regions, our method matches speech segments to textual labels.
Using a model that is more similar to ours, Kamper \etal~\cite{kamper2019a} investigated the task of retrieving utterances relevant to a given spoken query.
They did this by comparing the query embedding to the embedding of each of the utterance sub-segment using cosine distance, and then treated the minimum cosine distance as the score for the relevance of the query to the utterance. 
Here, we use their approach of splitting each utterance into overlapping segments varying from some minimum duration to some maximum duration.
But here the query is a written keyword (not a spoken query) and, much more importantly, \cite{kamper2019a} did not evaluate localisation performance.

    \subsubsection{Score aggregation}~
    \label{subsubsec:score-agg}
	
	The score aggregation method was originally designed in~\cite{palaz2016} to enable a model trained only for word classification to also locate a predicted word.
	It does so by explicitly making the architecture location-aware.
	The final convolutional layer must have the same number of filters as the final output of the model.
	By 
	aggregating the 
	output of this final convolutional layer in a specific way (described below), we can 
	 backtrace from the output prediction to determine 
	which of the filters %
	 at which time position 
	 causes that particular output prediction.

	More formally, the
    score aggregation method interprets the audio embeddings $\Hb = \left[\hb_1, \dots, \hb_\temp\right]$ as localisation scores $\alpha$, that is 
    $\alpha_{w, t} = h_{w, t}$.
    This method relies on two assumptions:
    (1) the embedding dimension $\emb$ is equal to the number of keywords in the vocabulary $\vocab$, and
    (2) that the pooling layer is the final layer, in order to preserve the semantics of the activations as scores. Hence, the method cannot be applied to an arbitrary network. 

    To aggregate the frame-level scores into an utterance-level detection score (across the $\temp$ steps), we use the log-mean-exp function
    followed by the sigmoid function $\sigma$ (which ensures that the values are in $[0, 1]$ and can be interpreted as probabilities):
    \begin{equation}
        \pool(\Hb) = \sigma\left(\frac{1}{r} \log \mean_{\temp} \exp \left(r \Hb\right)\right).
        \label{eq:log-mean-exp}
    \end{equation}
    Here $r$ is a positive number that controls the type of pooling:
    when $r \to 0$ we obtain average pooling,
    when $r \to \infty$ we obtain max pooling.

    \textbf{Relation to prior work.}
    Palaz \etal~\cite{palaz2016} proposed this score aggregation method to jointly locate and classify words using only BoW supervision.
    Here, we 
    train 
    the score aggregation method using visual supervision. 
    For image convolutional neural networks, a related technique is class activation mapping (CAM)~\cite{zhou2016},
    which decomposes the the classification score of a class as a sum of spatial scores from the last feature map in the convolutional neural network.
    This relies on the assumption that the last feature map is passed through a global average pooling layer and a linear classifier.
    This approach is related to Grad-CAM (Section~\ref{subsubsec:grad-cam}) since
        both methods compute spatial scores by pooling the convolutional feature maps across the channel axis.
        The main difference is
        that CAM aggregates using the linear classifier's weights
        while Grad-CAM uses gradients of the output with respect to the feature map.
    
    \begin{figure}[!t]
        \centering
        \includegraphics[width=\columnwidth]{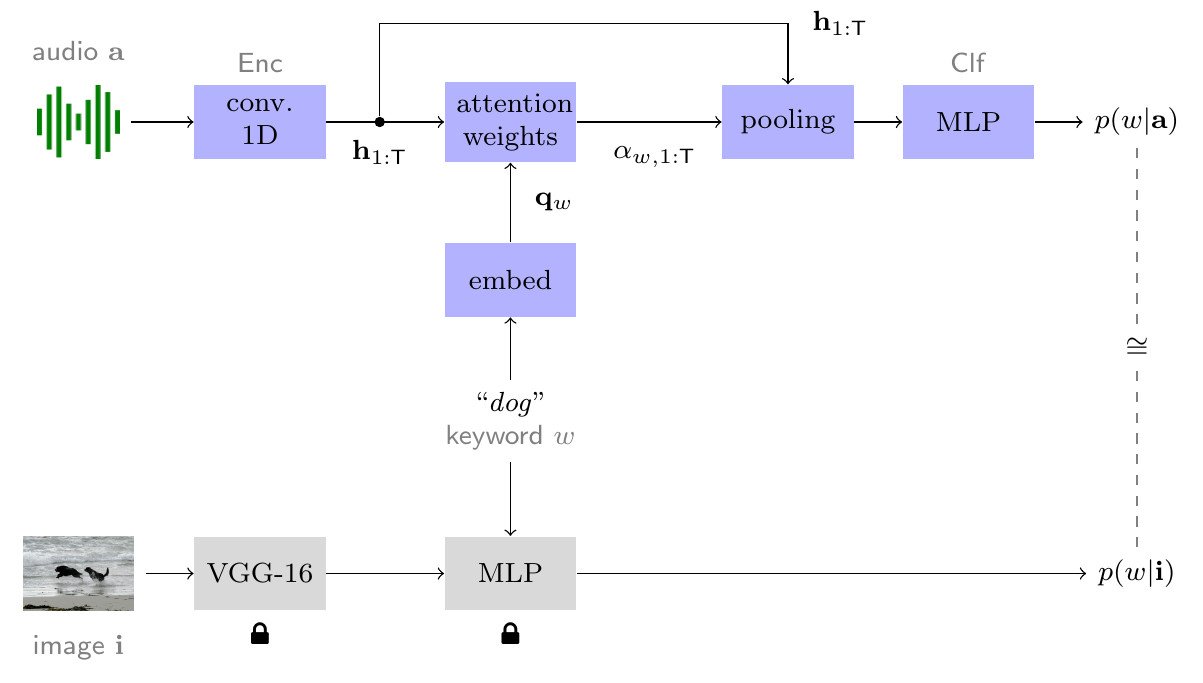}
        \caption{%
            Our attention-based architecture for keyword localisation in speech.
            The speech network (in blue) first passes the audio input $\ab$ through a convolutional encoder.
            The resulting features $\hb_{1:\mathsf{T}}$ are then pooled using an attention mechanism based on the embedding $\mathbf{q}_w$ of the query keyword $w$.
            The previous step yields a vector that finally goes through a multi-layer perceptron (MLP) to output the probability $p(w|\ab)$ of the keyword occurring in the audio.
            The visual network (in grey) embeds the corresponding image $\ib$ into a vector using the VGG-16 network and computes the probability $p(w|\ib)$ of the keyword occurring in the image
            using a keyword specific MLP.
            Note that the visual network is used only at training time (for supervising the speech branch) and its weights are fixed; at test time we only use the speech branch at the top.
        }
        \label{fig:attention_mechanism}
    \end{figure}

    \subsubsection{Attention}~
    \label{subsubsec:attention}

    The attention mechanism~\cite{graves2014,luong2015} performs localisation by using the input-dependent attention weights as localisation scores.
    Intuitively, a particular keyword is represented by a query embedding which is then used to weigh features over time. Time steps that are weighed highest (i.e.\ are attended to most) are likely to be most indicative of the location of a particular keyword.
    The method is illustrated in Figure~\ref{fig:attention_mechanism}.

    Concretely, we start by embedding the given keyword $w$ into a $\emb$-dimensional embedding $\mathbf{q}_w$ (query) using an index-based lookup table.
    The attention module then computes a score $e_{w, t}$ for each time step $t$ by using a dot product to measure the similarity between each time step along the convolutional features $\mathbf{h}_t$ and the query embedding,
    $e_{w, t} = \mathbf{q}_w^{\intercal} \mathbf{h}_t$,
    followed by a softmax operation which converts the similarity score $e_{w, t}$ to attention weights:
    \begin{equation}
        \alpha_{w, t} = \frac{\exp \{ e_{w, t} \} }{\sum_{{t' = 1}}^{\temp}\exp \{ e_{w, t'} \}}.
        \label{eq:attention-weights}
    \end{equation}
    The attention weights are used for both keyword localisation (as localisation scores) and keyword detection (as intermediary features).
    To obtain the detection score, we pool the features $\mathbf{h}_t$ at every time step $t$ into a keyword-specific context vector:
    \begin{equation}
        \pool(\Hb)_w = \sum_{t = 1}^{\temp} \alpha_{w, t} \mathbf{h}_t.
    \end{equation}
And, finally, we compute the score for query $w$ by passing the context vector through the classifier $\clf$,
    which is typically implemented as a series of fully-connected layers (a multi-layer perceptron).
  
    \textbf{Relation to prior work.}
    Tamer \etal~\cite{tamer2020} performed keyword search in sign language 
    using an attention-based graph-convolutional network over skeleton joints.
We adapted their architecture to fit our keyword localisation task in speech.
        The attention mechanism was also used for the general case of multiple instance learning~\cite{ilse2018}.
        This task assumes the classification of bags of instances:
        a bag is positive if contains at least one positive instance.
        The challenge lies in the fact that the labels are given only at the level of the bag, while the instance labels are unknown.
        The idea is to pool the instances using attention and use the corresponding weights as a indication of whether an instance is positive.
        As a practical use case, Ilse \etal \cite{ilse2018} applied this technique to highlight cancerous regions in weakly-labelled images.

\section{Experimental setup and evaluation}
\label{sec:exp_setup}

This section describes the data used in our experiments, the model architectures and how the models were implemented, and the tasks and metrics used to evaluate them.

\subsection{Data set}
\label{subsec:dataset}
\begin{figure}[bt]
		\centering
		\includegraphics[width=0.99\linewidth]{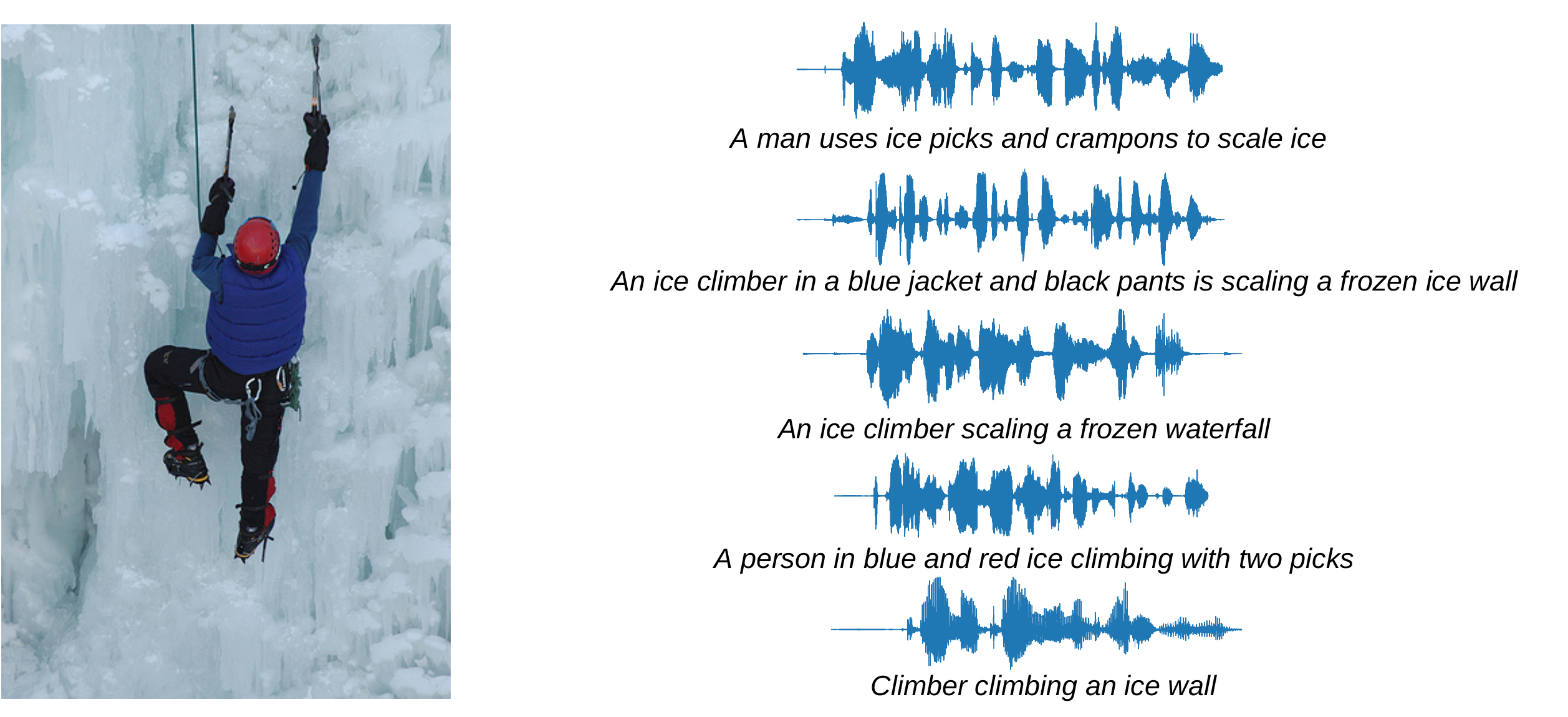}
        \caption{%
            An example from the Flickr8k Audio Caption Corpus~\cite{harwath2015} of an image with its corresponding spoken captions. The textual captions are not used during training and are given here only for illustrative purposes.
        }
		\label{fig:image_captions_examples}
\end{figure}
We perform experiments on the Flickr8k Audio Caption Corpus of~\cite{harwath2015}, consisting  of five English audio captions for each of the roughly $8$k images.
A split of $30$k, $5$k and $5$k  utterances are used for training, development and testing, respectively.
An example of an image with its corresponding spoken captions are shown in Figure~\ref{fig:image_captions_examples}.

The VGS models (Section~\ref{sec:vgs_models}) are trained using soft text labels (Section~\ref{subsec:methodological-overview}) obtained by passing each training image through the multi-label visual tagger of~\cite{kamper2019b}. This tagger 
is based on VGG-16~\cite{simonyan2014} and is trained on images disjoint from the data used here, specifically images from MSCOCO~\cite{lin2014}
This tagger has an output of 1000 image classes, but here we use a system vocabulary corresponding to $\vocab = 67$ unique keyword types.
This set of keywords is the same set used in~\cite{kamper2019a,kamper2019b}, and includes words such as \textit{children}, \textit{young}, \textit{swimming}, \textit{snowy} and \textit{riding}. 
The procedure  used to select these keywords are detailed in~\cite{kamper2019b}; it includes a human reviewer agreement step, which reduced the original set from 70 to 67 words.

\subsection{Model architectures and implementation details}
\label{subsec:model-architectures}

    We define our model 
    architectures in terms of the three components mentioned in Section~\ref{subsec:methodological-overview}:
    the encoder, pooling layers and the classifier.
    
\textbf{Encoders (Enc).}
We use two types of encoders: CNN and CNN-Pool.
Both are based on one-dimensional convolutional neural networks.
The main
difference between them is that CNN-Pool uses intermediate max-pooling while CNN doesn't. 
The reason we consider the two alternatives is that in
earlier studies,
CNN-Pool gave better scores on keyword detection while CNN gave better scores on keyword localisation in some settings~\cite{kamper2017a, olaleye2020, olaleye2021}.

The CNN-Pool encoder consists of three one-dimensional convolutional layers with ReLU activations.
Intermediate max-pooling over 3 units are applied in the first two layers. 
The first convolutional layer has 64 filters with a width of 9 frames. 
The second layer has 256 filters with a width of 11 units, and the last layer has 1024 filters with a width
of 11. 
The CNN encoder consists of 6 one-dimensional convolutional layers with ReLU activations. 
The first has 96 filters with a kernel width of 9 frames. 
The next four has a width of 11 units with 96 filters. 
The last convolutional layer uses 1000 filters with a width of 11 units.
A padding size of 4 is used
in the first convolutional layer, and 5 in the remaining layers.

\textbf{Pooling layers (Pool).}
All our models take in variable-duration speech data 
as input 
and then makes a fixed number of output predictions.
We therefore need to summarise 
the output from the speech encoder into a fixed-dimensional vector.
We specifically compare 
three types of pooling layers:
max pooling over time;
log-mean-exp pooling as described in Equation~\ref{eq:log-mean-exp}; and
attention pooling as described in Equation~\ref{eq:attention-weights}. 
The log-mean-exp pooling and attention pooling methods are specifically designed to give a model localisation capabilities.

\textbf{Classifiers (Clf).}
The classifier consists of a two-layer multi-layer perceptron (MLP) 
with rectified linear unit (ReLU) activations. The first layer produces an output of dimension 4096.
There are differences in terms of output dimensions of the second layer (precise details below): in some cases the classifier outputs a probability for each of the $\vocab$ words in the output vocabulary (e.g.\ when trained for multi-label classification), while other times it outputs a single value for whether a specific given keyword is present or not (e.g. when trained with attention).

    \textbf{Architectures.}
    Based on these components we assemble four types of architectures.
    These are summarised 
    in Table~\ref{tbl:architectures_localisation}.
    CNN, as used here, refers to the encoder type with no intermediate max pooling in its stack of convolutional layers (see above). CNN-Pool does include intermediate max pooling.
    We consider four different architectures which can be combined with four distinct localisation methods (as described in Section~\ref{subsec:methods-for-keyword-localisation}). 
    The PSC architecture uses log-mean-exp as its global pooling function, CNN as its audio encoder, and score aggregation as its localisation method (it gets its name from the author names of~\cite{palaz2016}).
    Some localisation methods are tied to particular architectures, e.g.\ score aggregation is directly incorporated into the PSC architecture and loss.
    Other localisation methods can be applied irrespective of the architecture, e.g.\ masking-based localisation can in principle be used with any architecture.
    The attention localisation method use a classifier that outputs a single detection score for whether a given keyword occurs.

    \begin{table}[t]
            \scriptsize
            \setlength{\tabcolsep}{3pt}
            \caption{Four architectures 
                in terms of their
                components (encoder, pooling layer, classifier) and
                localisation method (Grad-CAM \textsf{GC}, input masking \textsf{M}, score aggregation \textsf{SA}, attention \textsf{Att}).}
            \label{tbl:architectures_localisation}
            \centering
            \begin{tabular}{@{}llllccccc}
                \toprule
                Architecture       & $\enc$   & $\pool$      & $\clf$    &  & \textsf{GC} & \textsf{M} & \textsf{SA} & \textsf{Att} \\
                \midrule
                 PSC            & CNN      & log-mean-exp & ---       &  &             &            & $\bullet$   & \\
                 CNN-Pool       & CNN-Pool & max          & MLP(1024, $\vocab$) &  & $\bullet$   &            &             & \\
                 CNN-Attend     & CNN      & attention    & MLP(1000, 1) &  &             & $\bullet$  &             & $\bullet$ \\
                 CNN-PoolAttend & CNN-Pool & attention    & MLP(1000, 1) &  &             & $\bullet$  &             & $\bullet$ \\
                \bottomrule
            \end{tabular}
    \end{table}

\textbf{Implementation details.}
All architectures take as input mel-frequency cepstral coefficients. 
We augment the data using SpecAugment~\cite{park2019}, 
an augmentation policy that consists of warping the speech features, masking blocks of frequency channels, and masking blocks of time steps.
This policy achieves state-of-the-art performance on an end-to-end speech recognition tasks~\cite{park2019}. 
All models are implemented in PyTorch and use Adam optimisation~\cite{kingma2015} with a learning rate of $1\cdot{10}^{-4}$.
We train each model for 100 epochs and choose the best one based on its performance on the development set.
All models are trained on a single
GeForce RTX 2070 GPU with 8~Gb of RAM.
The code recipe to reproduce our 
results 
is available at \url{https://github.com/kayodeolaleye/keyword_localisation_speech}.

\subsection{Evaluation of localisation}
\label{subsec:evaluation-tasks}

Keyword localisation can be evaluated in different ways depending on the final use-case.
In some cases we might know definitely that a keyword occurs within an utterance and asked to predict where the keyword occurs (oracle keyword localisation, below).
In other cases we might first need to detect whether a keyword occurs before doing localisation (actual keyword localisation).
Or we might be required to rank utterances in a speech collection and only do localisation on the utterances that are most likely to contain the given keyword (keyword spotting localisation).
We give specific details below.

\textbf{Oracle keyword localisation.} 
In oracle keyword localisation, we have a scenario where we already know that a specified keyword occurs in an utterance and we need to find the location of that keyword. In this setup we do not need to first do detection since we already know that the keyword is present.
This scenario is referred to as \textit{oracle localisation performance} in~\cite{palaz2016}
and represents a setting where detection is perfect and we only consider keyword localisation performance.
Localisation is performed as in Equation~\ref{eq:location-max}. We calculate an accuracy: the proportion of proposed word locations that fall within the ground truth word boundaries.

\textbf{Actual keyword localisation.}
To perform \textit{actual keyword localisation}, we first check whether the detection score $y_w$ of keyword $w$ is greater than a specified threshold $\theta$,
    and then select the position of the highest attention weight $\alpha_{t, w}$ as the predicted location $\tau$ of $w$, as in Equation~\ref{eq:location-max}. 
    The location $\tau$ is accepted as correct if it falls within the interval corresponding to the true location of the word (which is determined based on the forced alignment between the text caption and audio utterance).
    The model is penalised whenever it localises a word whose detection score is less than or equal to $\theta$, i.e.\ if a word is not detected 
    then it is counted as a failed localisation even if $\alpha_{t, w}$ is at a maximum within that word.
    We set the value of $\theta$ to $0.5$ which gave best performance on development data.
    We report 
    precision, recall, and $F_1$.

\textbf{Keyword spotting localisation.}
    We adapt the related task of keyword spotting to the localisation scenario.
    In keyword spotting
    we are given a keyword and a collection of utterances and
    we want to rank the utterances according to the probability that the utterances contain the keyword.
     Keyword spotting is evaluated
    in terms of 
    $P@10$, the average precision of the ten highest-scoring proposals; and
    $P@N$, the average precision of the top $N$ proposals, with $N$ the number of true occurrences of the keyword. 
    In the case of \textit{keyword spotting localisation}, we additionally consider the localisation of the given keyword:
    a sample is deemed correct if it is both highly ranked (spotted) and the predicted location falls within the ground truth word.
	{Below we report keyword spotting localisation performance as $P@10$, which is the proportion of top-10 utterances for which the location is correctly predicted. If an utterance is erroneously ranked in the top 10 (i.e.\ the utterance doesn't contain the specified keyword), localisation performance is penalised (i.e.\ this utterance is considered a failed localisation).

\section{Experimental results and analysis}
\label{sec:exp_results}

We first answer our main research questions. We then turn to further quantitative and qualitative analyses. 

\subsection{Main findings}

\begin{table*}[!t]
\centering
\caption{%
    {Results on test data 
    for  three localisation tasks.
    We report results (in percentages) in terms of accuracy, $F_1$ score and precision at rank 10 ($P@10$), for the respective tasks.
    We consider both visual (VGS) and idealised bag-of-word (BoW) supervision.
    }
}
   \label{tbl:keyword_localisation_all_tasks_evaluation}
   \begin{tabularx}{1\linewidth}{@{}rllcRRRRRR}
    \toprule
    & & &
    & \multicolumn{2}{c}{Oracle (accuracy)}
    & \multicolumn{2}{c}{Actual ($F_1$)}
    & \multicolumn{2}{c}{Spotting ($P@10$)}
    \\
       \cmidrule(lr){5-6}
       \cmidrule(lr){7-8}
       \cmidrule(l){9-10}
                     & Architecture            & Localisation &  & VGS  & BoW  & VGS  & BoW & VGS & BoW \\
      \midrule
      \ii{1}         & {PSC}            & score agg.   &  & 13.7 & 61.5 & 7.4  & 68.1 & 8.2  & 62.2 \\
      \ii{2}         & {CNN-Pool}       & Grad-CAM     &  & 12.7 & 13.3 & 8.1  & 17.4 & 4.2  & 15.2 \\
      \ii{3}         & {CNN-Attend}     & attention    &  & 46.0 & 73.7 & \bf 25.2 & 72.1 & \bf 32.1 & 79.7 \\
      \ii{4}         & {CNN-Attend}     & masked-in    &  & \bf 57.3 & \bf 87.5 & 24.9 & \bf 79.8 & 21.9 & \bf 86.6\\
      \ii{5}         & {CNN-Attend}     & masked-out   &  & 25.1 & 28.2 & 12.9 & 57.8 & 5.1 & 7.9\\
      \ii{6}         & {CNN-Pool-Attend} & attention    &  & 16.5 & 33.6 & 8.2  & 39.0 & 8.1  & 42.7 \\
      \ii{7}         & {CNN-Pool-Attend} & masked-in    &  & 41.6 & 57.3 & 18.0 & 55.3 & 9.7 & 51.8\\
      \ii{8}         & {CNN-Pool-Attend} & masked-out   &  & 22.3 & 28.7 & 10.4 & 48.1 & 5.5 & 11.9\\
      \bottomrule
    \end{tabularx}
\end{table*}

Our main results are presented in Table~\ref{tbl:keyword_localisation_all_tasks_evaluation}.
We consider four different architectures which can be combined with four distinct localisation methods, as outlined in Table~\ref{tbl:architectures_localisation}.
Table~\ref{tbl:keyword_localisation_all_tasks_evaluation} shows the performance of both VGS 
models and BoW-supervised models on three tasks: oracle localisation, actual keyword localisation, and keyword spotting localisation.

Our main research goal is to investigate to what extent keyword localisation is possible with a VGS model.
Focussing therefore on the results in the ``VGS'' columns in
Table~\ref{tbl:keyword_localisation_all_tasks_evaluation}, we see that modest performance is achievable with visually supervised models.
In the oracle setting where the system knows that a keyword is present and then asked to predict where it occurs, accuracy is above 57\% using the best approach.
When localisation is performed after detection,
the best $F_1$ score is at 25.2\% while the best keyword spotting localisation $P@10$ is at 32.1\%.
Although these scores are modest, these keyword localisation results are achieved by models that do not receive any explicit textual supervision.

Of the architectures and localisation methods, which perform best when trained as VGS models?
Overall we see that the CNN-Attend and CNN-Pool-Attend architectures outperform the PSC and CNN-Pool architectures, and that the masked-in and attention localisation methods outperform the other localisation approaches.
This is the first time that masked-based localisation is incorporated with VGS modelling, and we see in oracle localisation that this approach outperforms its closest competitor by more than 10\% absolute in accuracy.
In contrast to the Grad-CAM and score aggregation localisation methods which are tied to a particular model structure, the masked-based localisation approaches have the benefit that they can be applied with any model architecture.
When using BoW supervision, best performance is always achieved with the CNN-Attend model employing masked-in prediction, while when using visual supervision, attention-based localisation works best on two of the three tasks. 
We speculate that the reason Grad-CAM performs poorly (Table~\ref{tbl:keyword_localisation_all_tasks_evaluation}, row 2) across all localisation tasks, regardless of the form of supervision, is because it tries to
identify the parts of the input that will cause a large change in a particular output unit: in single-label multi-class classification, for which Grad-CAM was developed, a higher probability for a particular output implies lower probabilities for others. But this is not the case for multi-label classification, as used here, and the gradients of multiple words could therefore affect the output, and this would be captured in the gradients. While the score aggregation method performs relatively well across all localisation tasks when BoW supervision is available during training, we observe that the performances drops 
when we move to visual supervision (Table~\ref{tbl:keyword_localisation_all_tasks_evaluation}, row 1).

What is the penalty we pay in using visual supervision instead of BoW supervision?
This is an important question since BoW supervision can be seen as a setting where we have access to a perfect visual tagger:
BoW labels are still weak (they do not contain any location information), but all the words in the bag are related to the image (according to a caption provided by a human).
By comparing the ``VGS'' and ``BoW'' columns in Table~\ref{tbl:keyword_localisation_all_tasks_evaluation}, we see that in all cases there is a large drop in performance when moving from idealised supervision to visual supervision. For instance, for actual localisation, the best approach drops from around 80\% to around 25\% in $F_1$.
Improving visual tagging could therefore lead to substantial improvements for keyword localisation using VGS models.
It should be noted, however, that although the BoW does give an upper bound for VGS performance, this could be a high upper bound: a perfect visual tagger might detect objects that aren't described by humans, and in other cases it might not have visual support for a keyword that might occur in an utterance (e.g.\ \textit{camera} in ``a girl is posing in front of the camera'').

As a sub-research question, we note that improving keyword localisation for BoW-supervised models is in itself a useful endeavour. 
BoW labels can be easier to obtain than transcriptions in many cases, see e.g.~\cite{van2022}.
Localisation with BoW supervision was also the main goal of the original PSC study~\cite{palaz2016}.
Comparing the BoW models, we see that our CNN-Attend model with masked-in localisation gives a substantial improvement over the PSC approach on all three tasks, with absolute improvements in accuracy and $F_1$ scores of more than 10\% absolute in all three cases.

\subsection{Further quantitative analysis}

We now turn to further analyses: we try to quantify the drop in performance due to the first detection or ranking pass, we consider the effect of architectural choices, and break down the performance for individual keywords.

\begin{table}[!t]
	\centering
	\caption{{
		Actual keyword localisation precision, recall and $F_1$ scores for VGS models.
		Keyword detection scores are given in parentheses; since actual localisation involves first doing detection and only then locating words, the detection scores serve as upper bounds on localisation performance.
		}}
	\label{tbl:actual_upperbound}
	\begin{tabular}{rllccc}
		\toprule
		                        & Architecture                       & Localisation               & $P$    & $R$    & $F_1$  \\
        \midrule
        \multirow{2}{*}{\ii{3}} & \multirow{2}{*}{CNN-Attend}        & \multirow{2}{*}{attention} & 28.8   & 22.4   & 25.2   \\
                                &                                    &                            & (38.9) & (28.2) & (32.7) \\[3pt]
        \multirow{2}{*}{\ii{4}} & \multirow{2}{*}{{CNN-Attend}}      & \multirow{2}{*}{masked-in} & 28.3   & 22.2   & 24.9   \\
                                &                                    &                            & (38.9) & (28.2) & (32.7) \\[3pt]
        \multirow{2}{*}{\ii{7}} & \multirow{2}{*}{{CNN-Pool-Attend}} & \multirow{2}{*}{masked-in} & 21.9   & 15.3   & 18.0   \\
		                        &                                    &                            & (35.1) & (22.6) & (27.5) \\
		\bottomrule
	\end{tabular}
\end{table}

\textbf{Limits on localisation performance from first-pass detection.}
When doing actual keyword localisation (middle columns in Table~\ref{tbl:keyword_localisation_all_tasks_evaluation}), we are limited by the first detection pass which only considers keywords above a threshold ($\theta = 0.5$).
Since we can't locate words that aren't correctly detected, the detection scores are upper bounds on the localisation performance that can be achieved.

To quantify the limits on localisation performance for a subset of the VGS models, Table~\ref{tbl:actual_upperbound} gives the recall, precision and $F_1$ scores that were achieved together wit the keyword detection scores (given in parentheses).
(One way to see the scores in parentheses is in analogy to the oracle results in Table~\ref{tbl:keyword_localisation_all_tasks_evaluation}: for oracle localisation we assumed that detection was perfect and only considered localisation; now we consider localisation to be perfect and consider detection performance.)
We see in all cases that localisation is impaired by the first detection pass. For recall, we are within roughly 6\% of the upper-bound due to detection errors, while for precision we are roughly within 10\%.
Future work could consider an approach where detection and localisation are performed using different models tailored to these distinct tasks.

\textbf{Limits on localisation from first-pass keyword spotting.}
In a similar way, the keyword spotting localisation results (right, Table~\ref{tbl:keyword_localisation_all_tasks_evaluation}) are limited by the first ranking pass as performed in standard keyword spotting.
I.e., it is not possible to correctly locate a keyword if the first pass incorrectly ranked an utterance at the top but the utterance doesn't actually contain the given keyword.
Here Table~\ref{tbl:spotting_upperbound} shows the achieved $P@10$ localisation scores together with the upper-bound keyword spotting $P@10$ (in parentheses) for a subset of the VGS models.
We also include the $P@N$ scores here: achieved (and upper-bound).
We again see that localisation performance is limited: even for the best approach (CNN-Attend with attention) only about 10\% and 5\% of $P@10$ and $P@N$ localisation mistakes are not the cause of ranking errors.
Improvements in first-pass keyword spotting could therefore again potentially improve subsequent localisation.

\begin{table}[!t]
	\centering
	\caption{{
		Keyword spotting localisation scores in terms of $P@10$ and $P@N$ for VGS models.
		The corresponding keyword spotting scores (ignoring localisation) are given in parentheses; these are upper bounds on localisation performance.
	}}
	\label{tbl:spotting_upperbound}
	\begin{tabularx}{1\linewidth}{@{}rllCC}
		\toprule
		& Architecture & Localisation & $P@10$ & $P@N$ \\
		\midrule
		\ii{3} & {CNN-Attend}      & attention & 32.1 (43.7) & 24.9 (30.1) \\
		\ii{4} & {CNN-Attend}      & masked-in & 21.9 (43.7) & 19.5 (30.1) \\
		\ii{7} & {CNN-Pool-Attend} & masked-in & \hphantom{0}9.7 (31.6) & 10.4 (24.4)  \\
		\bottomrule
	\end{tabularx}
\end{table}

\textbf{Effect of intermediate max pooling on localisation performance.}
Another way to analyse the different architectures and localisation methods is by looking at how their choice of encoder structure affect their localisation capabilities. Each architecture either uses CNN or CNN-Pool. In the CNN-Pool encoder, each convolutional layer operation (except the last one) is followed by a max pooling operation. The CNN encoder type doesn't perform such intermediate pooling operations. Intuitively, the max pooling operation calculates and returns the most prominent feature of its input by downsampling along the time axis. While such operation has been shown to help with better word detection~\cite{kamper2017a}, it hurts localisation performance in both the VGS and BoW supervision settings. We demonstrate this by comparing localisation scores of our two best localisation methods for the two encoder types.
As shown in Table~\ref{tbl:keyword_localisation_all_tasks_evaluation}, %
all architectures with the CNN encoder type that uses the attention and masked-in localisation methods outperforms the ones with the CNN-Pool encoder type on the three tasks. This behaviour is expected since the max pooling operation employed by the CNN-Pool makes it more challenging to backtrace each output prediction to its most relevant filter.

\begin{figure}
    \centering
    \includegraphics[width=0.95\linewidth]{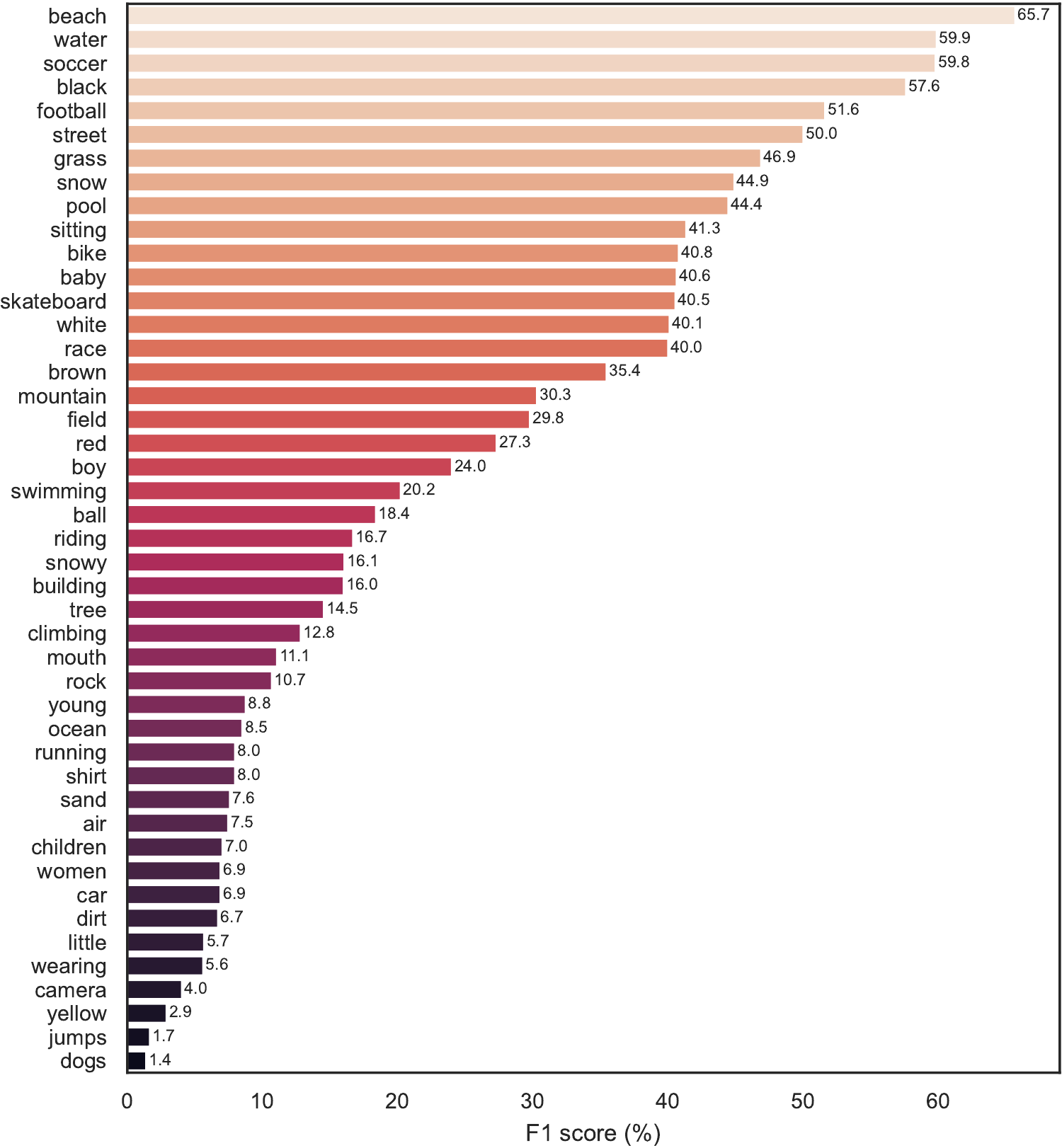}
    \caption{%
        Actual localisation performance measured in terms of $F_1$ score reported per keyword.
        We include only those keywords (45 out of 67) for which at least one utterance was detected, that is, the detection score is greater than the threshold $\theta$ of 0.5).
        The results are reported for the CNN-Attend architecture and the masked-in localisation method.
    }
    \label{fig:keyword-evaluation-f1}
\end{figure}


    \textbf{Per keyword performance.}
    We investigate the localisation performance for each individual keyword in the vocabulary.
    We report the results in the actual keyword localisation setup, that is,
    for each keyword $w$, we first select the utterances that have a detection score over the threshold $\theta$ and
    then check if the localisation is correct by verifying whether the location $\tau_w$ of the maximum localisation score falls into a segment corresponding to the query keyword $w$.
    The results are presented in Figure \ref{fig:keyword-evaluation-f1}.

    We report performance for 45 out of the total 67 keywords in the vocabulary, since for the remaining 22 keywords no utterance had a detection score over the imposed threshold $\theta$ of 0.5;
    when no utterance is detected we cannot localise it---the precision is not defined, while the recall is 0.
    We observe that the actual localisation performance varies strongly across the keywords in the vocabulary (from 1.4\% for \textit{dogs} to 65.7\% for \textit{beach}) suggesting that the methods are very keyword-dependent.
    Our assumption is that the keyword performance is influenced by how well each word is visually grounded.

    To test this hypothesis we look at how well the localisation performance correlates with the performance of the visual network; see Figure \ref{fig:keyword-evaluation-loc-vs-vis}.
    We observe that indeed there is a dependency between the two systems:
    keywords like \textit{camera}, \textit{air}, and \textit{wearing} understandably have poor visual grounding, which impacts the speech localisation capability;
    while keywords like \textit{beach}, \textit{soccer}, \textit{snow} are easily spotted by the visual network and also by the speech-based localisation network.
    However, there are some outliers: the keywords \textit{football}, \textit{car}, \textit{dogs} while well visually grounded, suffer when localised.
    We investigate these cases in Section~\ref{ssec:qualitative-results} and observe that they are caused by co-occurrences of the keywords with other words: \textit{football}--\textit{player}, \textit{car}--\textit{race}, \textit{dogs}--\textit{three} (see Table \ref{tbl:most-common-confusions}).
    Words that often occur together in the same utterance are very challenging to be discriminated by any localisation method that does not have access to additional information.

    \begin{figure}
        \centering
        \includegraphics[width=0.95\linewidth]{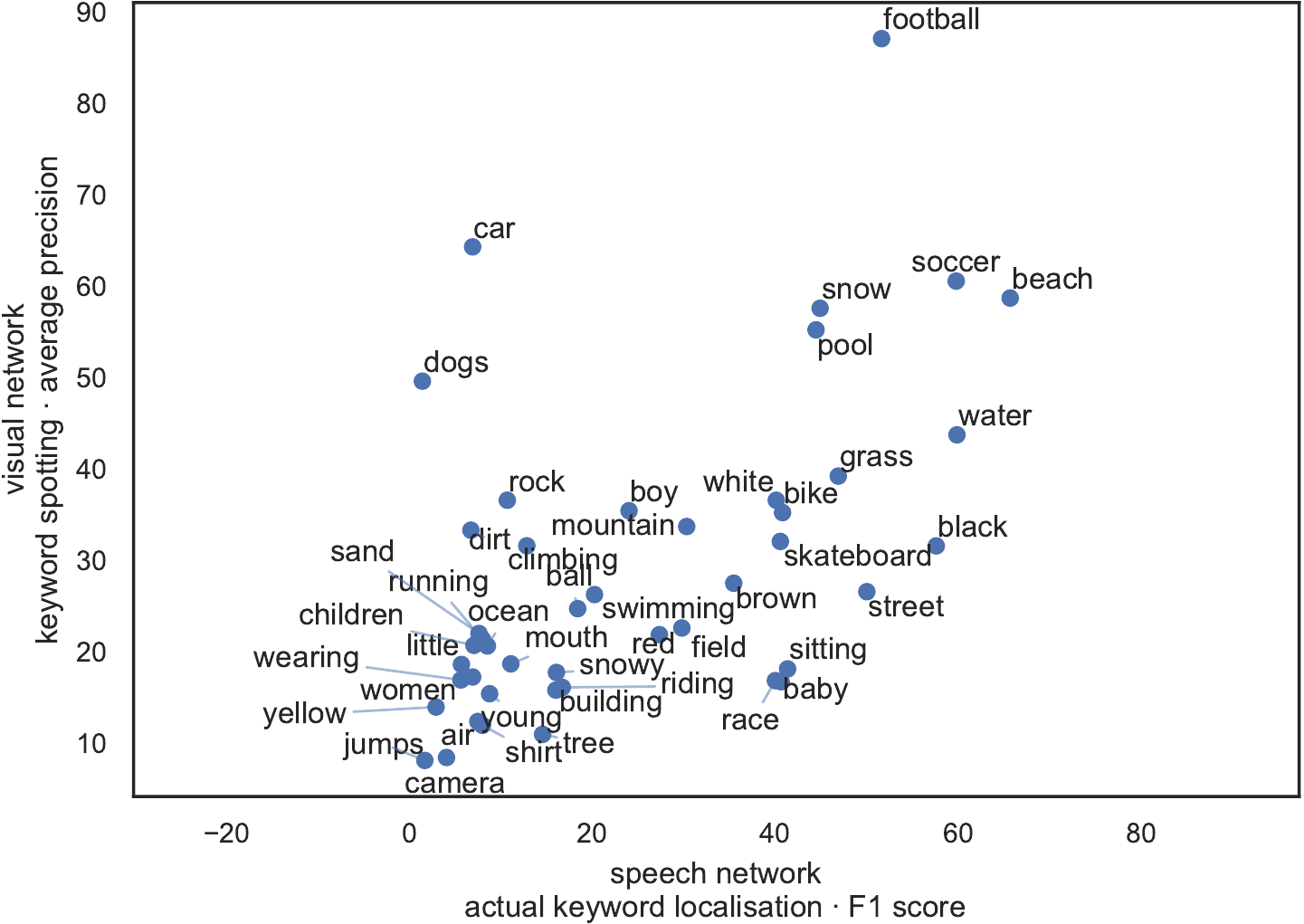}
        \caption{%
                Localisation performance ($F_1$) of the speech network versus spotting performance (average precision) of the visual network for each of the 44 keywords that were included in Figure \ref{fig:keyword-evaluation-f1}.
                The results are reported for the CNN-Attend architecture and the masked-in localisation method.
            }
        \label{fig:keyword-evaluation-loc-vs-vis}
    \end{figure}

    \vspace{0.5cm}

\subsection{Qualitative results}
\label{ssec:qualitative-results}

In this subsection we report concrete examples obtained by manually inspecting the samples and the network's predictions.

\begin{figure}[t]
        
        \centering
        \includegraphics[height=2.65cm]{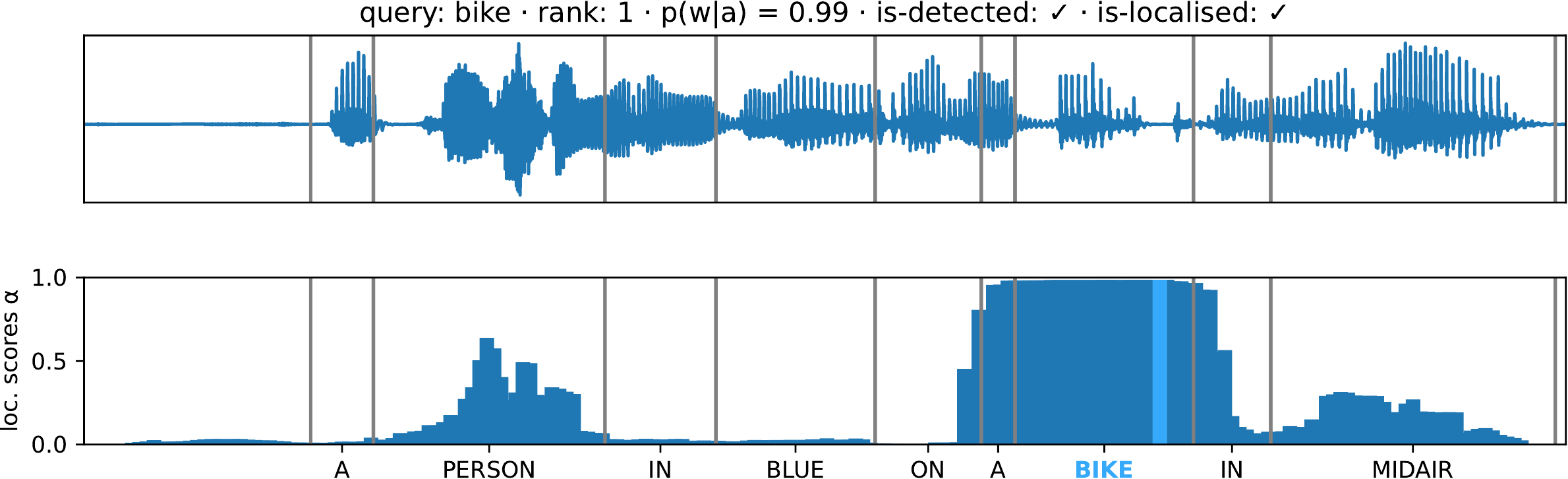} \\[7.5pt]
        \includegraphics[height=2.65cm]{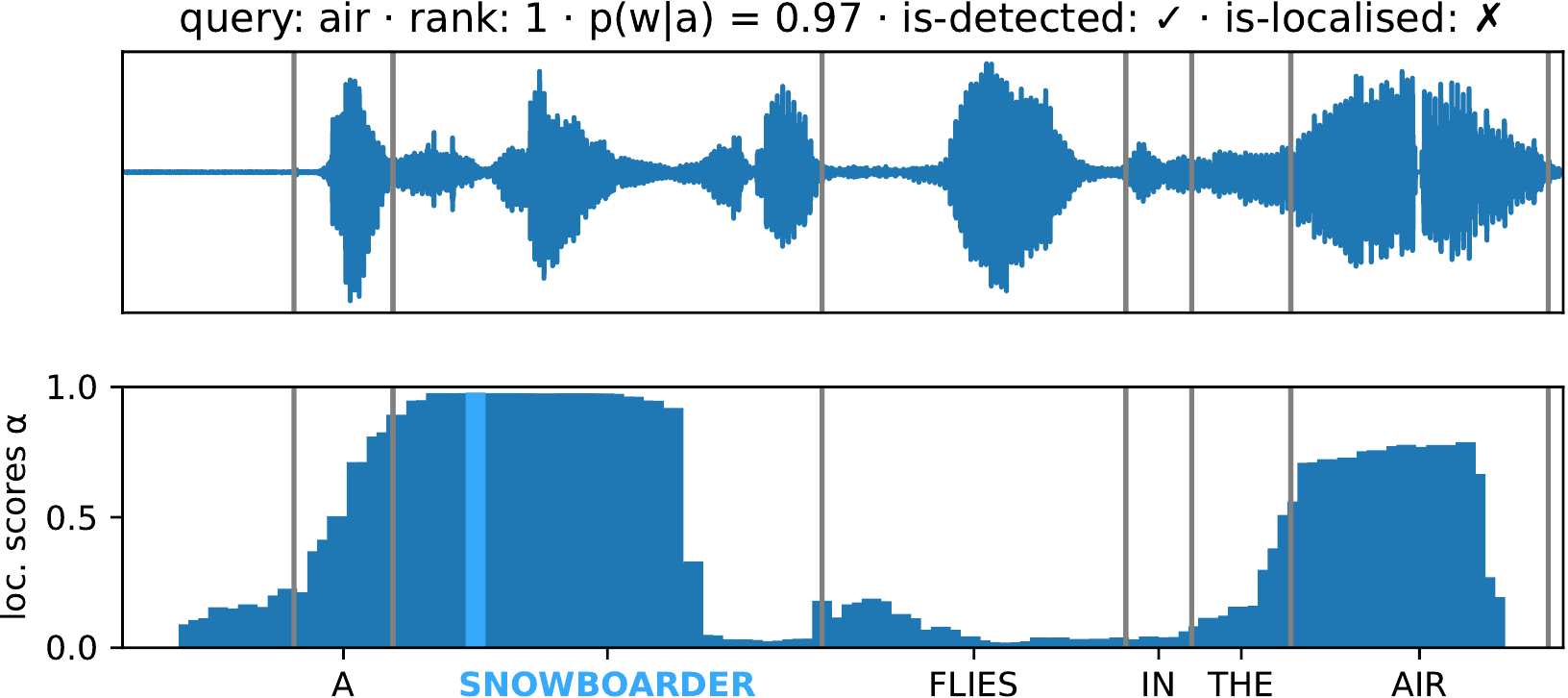} \\[7.5pt]
        \includegraphics[height=1.65cm]{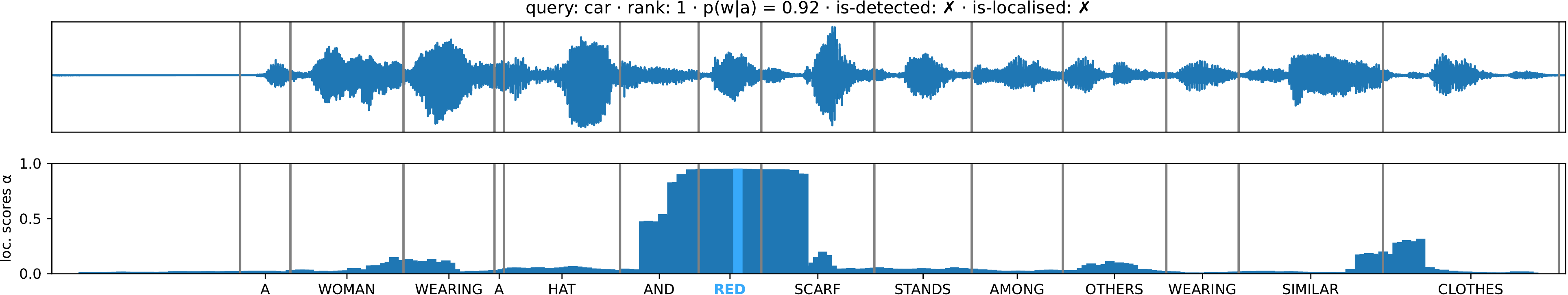} \\[7.5pt]
        \includegraphics[height=2.1cm]{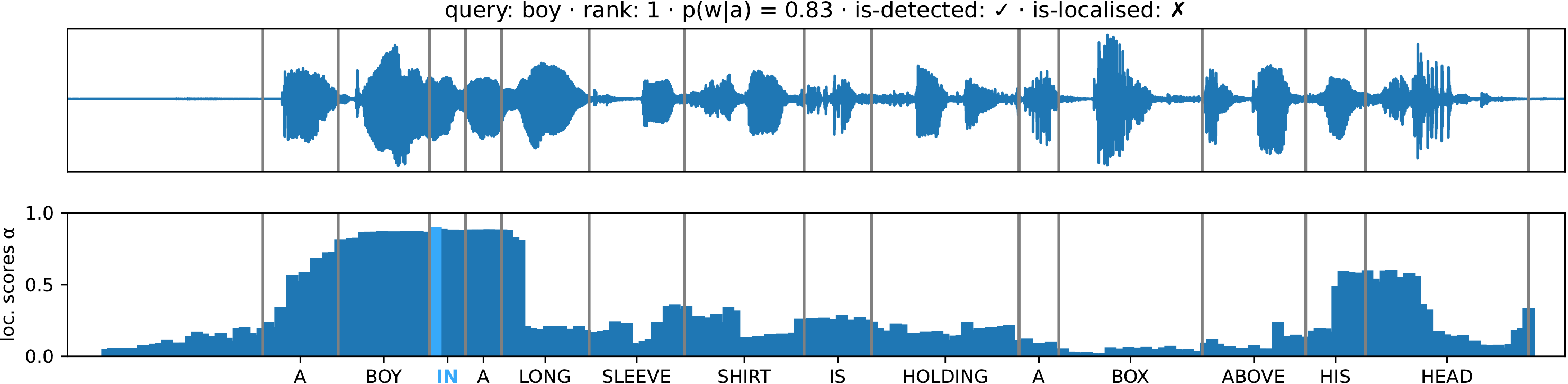}
        \caption{%
            Examples of predictions for four query keywords (\textit{bike}, \textit{air}, \textit{car}, \textit{boy}) using the CNN-Attend architecture and the masked-in localisation method.
            In each figure we show the input audio signal (top subplot) and the output localisation scores (bottom subplot).
            We also provide the detection score $p(w|\ab)$, whether each keyword was correctly detected or localised, the transcription of the utterance.
            The vertical grey lines denote the word boundaries and the word coloured in light blue shows the top scoring word.\label{fig:qualitative-samples}
        }
\end{figure}

\textbf{Qualitative samples.}
In Figure \ref{fig:qualitative-samples} we show the top-ranked utterances (based on their detection scores) and their corresponding localisation scores for four keywords:
\textit{bike}, \textit{air}, \textit{car}, \textit{boy}.
For \textit{bike}, we see that both detection and localisation are correct,
both achieving high scores close to the maximum value of 1.
In the case of the second example where the query keyword is \textit{air},
the top-ranked utterance does contain the query (hence the detection is correct),
but the word \textit{snowboarder} produces a higher localisation score than \textit{air}, yielding an incorrect localisation.
As we will see in the next set of experiments, \textit{air} is often confused with \textit{snowboarder} due to the fact they frequently co-occur.
The third example highlights another common confusion---\textit{red} is often predicted when the query is \textit{car}---%
in this case probably exacerbated by the phonetic similarity between \textit{red scarf} and the more common phrase \textit{red car}.
The final example shows a case of a narrowly missed localisation: while the word \textit{boy} achieves a large score, the maximum value is obtained at the boundary with the word \textit{in}.
This incorrect localisation could be due to the inherent imprecision of the forced alignment procedure, which is used to assign location information to the words in the transcription.

\textbf{Most common confusions.}
For each keyword in the vocabulary, we find which words in the audio utterance they are most commonly confused with.
To generate the results, we take the following steps.
(1) Given a keyword, we sort the utterances in decreasing order of their detection (per-utterance) score.
(2) For the each of the top twenty utterances, we choose the predicted word as the word that contains the maximum of the localisation scores $\alpha_{w,1}, \dots, \alpha_{w,\temp}$.
(3) We report the top five most commonly predicted words together with their counts.
Table \ref{tbl:most-common-confusions} shows the results for a subset of the keywords.

\begin{table}
        \scriptsize
        \setlength{\tabcolsep}{3pt}
		\centering
		\caption{%
			The most commonly located words for a subset of sixteen query keywords.
			For each keyword, we consider the top 20 utterances based on the detection scores and in each utterance we find the predicted word based on the localisation scores.
			We report the most common 5 located words.
            We use the CNN-Attend architecture and the masked-in localisation method.
			The 
            `--' symbol denotes spoken noise or words that are missing from the phonetic dictionary of the forced aligner.
		}
		\label{tbl:most-common-confusions}
		\begin{tabular}{@{}llrlrlrlrlr}
			\toprule
			Query & \multicolumn{10}{l}{Top 5 located words and their counts} \\
			\midrule
			air        & snowboarder & 11 & bike        & 2 & --            & 2 & snowboard   & 1 & skiing      & 1 \\
			ball       & soccer      &  9 & ball        & 5 & player        & 2 & basketball  & 2 & colorful    & 1 \\
			black      & black       & 18 & dog         & 2 &               &   &             &   &             &   \\
			car        & race        &  7 & red         & 4 & racetrack     & 2 & raises      & 2 & car         & 2 \\
			dogs       & three       &  8 & two         & 6 & dog           & 2 & puppy       & 1 & dogs        & 1 \\
			face       & climbing    &  7 & rock        & 5 & goes          & 1 & rocks       & 1 & blonde      & 1 \\
			football   & football    & 15 & player      & 4 & players       & 1 &             &   &             &   \\
			ocean      & surfer      & 11 & surfboard   & 4 & wave          & 3 & surfing     & 1 & surface     & 1 \\
			orange     & red         &  9 & wears       & 1 & a             & 1 & retriever   & 1 & guitar      & 1 \\
			pink       & girl        &  8 & young       & 2 & girls         & 2 & dressed     & 1 & one         & 1 \\
			pool       & pool        & 18 & swimming    & 1 & swimmer       & 1 &             &   &             &   \\
			soccer     & soccer      & 20 &             &   &               &   &             &   &             &   \\
			swimming   & pool        & 16 & swimming    & 3 & swim          & 1 &             &   &             &   \\
			tree       & tree        &  9 & trees       & 7 & training      & 1 & climbs      & 1 & --          & 1 \\
			yellow     & yellow      & 17 & bikes       & 1 & bike          & 1 & race        & 1 &             &   \\
			\bottomrule
		\end{tabular}
\end{table}

We observe good matches for some keywords (\textit{black}, \textit{pool}, \textit{soccer}, \textit{tree}),
while others are confused with semantically related words:
\textit{air} $\to$ \textit{snowboarder}; \textit{ocean} $\to$ \textit{surfer}; \textit{ball} $\to$ \textit{soccer}; \textit{swimming} $\to$ \textit{pool}; \textit{face} $\to$ \textit{climbing}, \textit{rock}.
The performance for \textit{car} is surprising: given the fact that \textit{car} has visual support, we would have expected better localisation, instead the word is often confused with \textit{race} and \textit{red}.
Interestingly, the keyword \textit{dogs} is sometimes associated with numerals (\textit{two} and \textit{three});
we believe this happens because we distinguish between singular and plural words (the word \textit{dog} is different from the word \textit{dogs}).
For some keywords representing colors (\textit{black}, \textit{white}, \textit{yellow}) we obtain strong results,
while for others there are more semantic confusions: \textit{pink} $\to$ \textit{girl}; \textit{orange} $\to$ \textit{red}.


\section{Conclusions}
\label{sec:conclusions}

We considered different approaches for doing keyword localisation using visually grounded speech (VGS) models that learn from images and unlabelled spoken captions.
We specifically used a VGS method that tags training images with soft textual labels as targets for a speech network that can then detect the presence of a keyword in an utterance.
We equipped this type of VGS model with localisation capabilities, investigating four localisation approaches.
Our best approach relied on masking (an approach not considered before): only a portion of the spoken input is passed through the model, and by looking at the change in output probability, a decision can be made regarding the presence of a keyword within the unmasked region of the utterance.
By using a sliding window with a changing width, the best location for a keyword can be identified.
This approach gave a 57\% localisation accuracy in a setting where we assume the system knows that a keyword occurs in an utterance and need to predict its location and an $F_1$ of 25\% when detection is first performed.
In further analyses we showed that system performance is limited by the first detection or ranking pass. Large performance could be possible by improving the visual tagger since per-keyword performance are strongly correlated with the performance of the visual tagger on that keyword. This will be the focus of future work.
We also showed that many incorrect localisations is because semantically related words are located---seeing what benefit this holds is another avenue for future work.

\bibliography{refs}

\begin{thebibliography}{10}
\providecommand{\url}[1]{#1}
\csname url@samestyle\endcsname
\providecommand{\newblock}{\relax}
\providecommand{\bibinfo}[2]{#2}
\providecommand{\BIBentrySTDinterwordspacing}{\spaceskip=0pt\relax}
\providecommand{\BIBentryALTinterwordstretchfactor}{4}
\providecommand{\BIBentryALTinterwordspacing}{\spaceskip=\fontdimen2\font plus
\BIBentryALTinterwordstretchfactor\fontdimen3\font minus
  \fontdimen4\font\relax}
\providecommand{\BIBforeignlanguage}[2]{{%
\expandafter\ifx\csname l@#1\endcsname\relax
\typeout{** WARNING: IEEEtran.bst: No hyphenation pattern has been}%
\typeout{** loaded for the language `#1'. Using the pattern for}%
\typeout{** the default language instead.}%
\else
\language=\csname l@#1\endcsname
\fi
#2}}
\providecommand{\BIBdecl}{\relax}
\BIBdecl

\bibitem{duong2016}
L.~Duong, A.~Anastasopoulos, D.~Chiang, S.~Bird, and T.~Cohn, ``An attentional
  model for speech translation without transcription,'' in \emph{Proc.
  NAACL-HLT}, 2016.

\bibitem{palaz2016}
D.~Palaz, G.~Synnaeve, and R.~Collobert, ``Jointly learning to locate and
  classify words using convolutional networks,'' in \emph{Proc. Interspeech},
  2016.

\bibitem{settle2017}
S.~Settle, K.~Levin, H.~Kamper, and K.~Livescu, ``Query-by-example search with
  discriminative neural acoustic word embeddings,'' in \emph{Proc.
  Interspeech}, 2017.

\bibitem{weiss2017}
R.~J. Weiss, J.~Chorowski, N.~Jaitly, Y.~Wu, and Z.~Chen,
  ``Sequence-to-sequence models can directly translate foreign speech,'' in
  \emph{Proc. Interspeech}, 2017.

\bibitem{driesen2010}
J.~Driesen, H.~Van~hamme, and W.~B. Kleijn, ``Learning from images and speech
  with non-negative matrix factorization enhanced by input space scaling,'' in
  \emph{Proc. SLT}, 2010.

\bibitem{synnaeve2014b}
G.~Synnaeve, M.~Versteegh, and E.~Dupoux, ``Learning words from images and
  speech,'' \emph{NeurIPS Workshop Learn}, 2014.

\bibitem{harwath2015}
D.~Harwath and J.~Glass, ``Deep multimodal semantic embeddings for speech and
  images,'' in \emph{Proc. ASRU}, 2015.

\bibitem{harwath2016}
D.~Harwath, A.~Torralba, and J.~Glass, ``Unsupervised learning of spoken
  language with visual context,'' in \emph{Proc. NeurIPS}, 2016.

\bibitem{harwath2017}
D.~Harwath and J.~Glass, ``Learning word-like units from joint audio-visual
  analysis,'' in \emph{Proc. ACL}, 2017.

\bibitem{harwath2018a}
D.~Harwath, A.~Recasens, D.~Sur{\'\i}s, G.~Chuang, A.~Torralba, and J.~Glass,
  ``Jointly discovering visual objects and spoken words from raw sensory
  input,'' in \emph{Proc. ECCV}, 2018.

\bibitem{harwath2018b}
D.~Harwath, G.~Chuang, and J.~Glass, ``Vision as an interlingua: Learning
  multilingual semantic embeddings of untranscribed speech,'' in \emph{Proc.
  ICASSP}, 2018.

\bibitem{eloff2019}
R.~Eloff, H.~A. Engelbrecht, and H.~Kamper, ``Multimodal one-shot learning of
  speech and images,'' in \emph{Proc. ICASSP}, 2019.

\bibitem{harwath2019a}
D.~Harwath and J.~Glass, ``Towards visually grounded sub-word speech unit
  discovery,'' in \emph{Proc. ICASSP}, 2019.

\bibitem{harwath2019b}
D.~Harwath, W.-N. Hsu, and J.~Glass, ``Learning hierarchical discrete
  linguistic units from visually-grounded speech,'' in \emph{Proc. ICLR}, 2019.

\bibitem{bomba1983}
P.~C. Bomba and E.~R. Siqueland, ``The nature and structure of infant form
  categories,'' \emph{J. Exp. Child Psychol.}, 1983.

\bibitem{pinker1994}
S.~Pinker, ``The language instinct,'' \emph{Harper Perennial, New York}, 1994.

\bibitem{eimas+quinn94}
P.~D. Eimas and P.~C. Quinn, ``Studies on the formation of perceptually based
  basic-level categories in young infants,'' \emph{Child Develop.}, 1994.

\bibitem{roy2003}
D.~Roy, ``Grounded spoken language acquisition: experiments in word learning,''
  \emph{IEEE Trans. Multimedia}, 2003.

\bibitem{boves2007}
L.~Boves, L.~ten Bosch, and R.~Moore, ``{ACORNS} -- towards computational
  modeling of communication and recognition skills,'' in \emph{Proc. ICCI},
  2007.

\bibitem{chrupala2016}
L.~Gelderloos and G.~Chrupa{\l}a, ``From phonemes to images: levels of
  representation in a recurrent neural model of visually-grounded language
  learning,'' in \emph{Proc. COLING}, 2016.

\bibitem{okko2019}
O.~R{\"{a}}s{\"{a}}nen and K.~Khorrami, ``A computational model of early
  language acquisition from audiovisual experiences of young infants,'' in
  \emph{Proc. Interspeech}, 2019.

\bibitem{meng2013}
M.~Sun and H.~{Van hamme}, ``Joint training of non-negative tucker
  decomposition and discrete density hidden markov models,'' \emph{Comput.
  Speech Lang.}, 2013.

\bibitem{nortje2020}
L.~Nortje and H.~Kamper, ``Unsupervised vs. transfer learning for multimodal
  one-shot matching of speech and images,'' in \emph{Proc. Interspeech}, 2020.

\bibitem{de1998}
V.~De~Sa and D.~Ballard, ``Category learning through multimodality sensing,''
  \emph{Neural Comput.}, 1998.

\bibitem{scharenborg2018}
O.~Scharenborg, L.~Besacier, A.~Black, M.~Hasegawa-Johnson, F.~Metze,
  G.~Neubig, S.~Stüker, P.~Godard, M.~Müller, L.~Ondel, S.~Palaskar,
  P.~Arthur, F.~Ciannella, M.~Du, E.~Larsen, D.~Merkx, R.~Riad, L.~Wang, and
  E.~Dupoux, ``Linguistic unit discovery from multi-modal inputs in unwritten
  languages: Summary of the “speaking rosetta” jsalt 2017 workshop,'' in
  \emph{Proc. ICASSP}, 2018.

\bibitem{lupke2010}
F.~L{\"u}pke, ``Research methods in language documentation,'' \emph{Lang.
  Documentation and Description}, 2010.

\bibitem{bird2020}
S.~Bird, ``Sparse transcription,'' \emph{Comput. Linguistics}, 2020.

\bibitem{scharenborg2020}
O.~Scharenborg, L.~Besacier, A.~Black, M.~Hasegawa-Johnson, F.~Metze,
  G.~Neubig, S.~Stüker, P.~Godard, M.~Müller, L.~Ondel, S.~Palaskar,
  P.~Arthur, F.~Ciannella, M.~Du, E.~Larsen, D.~Merkx, R.~Riad, L.~Wang, and
  E.~Dupoux, ``Speech technology for unwritten languages,'' \emph{IEEE/ACM
  TASLP}, 2020.

\bibitem{wang2021a}
X.~Wang, T.~Qiao, J.~Zhu, A.~Hanjalic, and O.~Scharenborg, ``Generating images
  from spoken descriptions,'' \emph{IEEE/ACM TASLP}, 2021.

\bibitem{tsai2021}
Y.-H.~H. Tsai, Y.~Wu, R.~Salakhutdinov, and L.-P. Morency, ``Self-supervised
  learning from a multi-view perspective,'' in \emph{Proc. ICLR}, 2021.

\bibitem{chrupala2017}
G.~Chrupa{\l}a, L.~Gelderloos, and A.~Alishahi, ``Representations of language
  in a model of visually grounded speech signal,'' in \emph{Proc. ACL}, 2017.

\bibitem{havard2019a}
W.~Havard, J.~Chevrot, and L.~Besacier, ``Models of visually grounded speech
  signal pay attention to nouns: a bilingual experiment on english and
  japanese,'' in \emph{Proc. ICASSP}, 2019.

\bibitem{havard2019b}
------, ``Word recognition, competition, and activation in a model of visually
  grounded speech,'' in \emph{Proc. CoNLL}, 2019.

\bibitem{kamper2017a}
H.~Kamper, S.~Settle, G.~Shakhnarovich, and K.~Livescu, ``Visually grounded
  learning of keyword prediction from untranscribed speech,'' in \emph{Proc.
  Interspeech}, 2017.

\bibitem{kamper2019b}
H.~Kamper, G.~Shakhnarovich, and K.~Livescu, ``Semantic speech retrieval with a
  visually grounded model of untranscribed speech,'' \emph{IEEE/ACM Trans.
  Acoust., Speech, Signal Process.}, 2019.

\bibitem{pasad2019}
A.~Pasad, B.~Shi, H.~Kamper, and K.~Livescu, ``On the contributions of visual
  and textual supervision in low-resource semantic speech retrieval,'' in
  \emph{Proc. Interspeech}, 2019.

\bibitem{imagenet2009}
J.~Deng, W.~Dong, R.~Socher, L.-J. Li, K.~Li, and L.~Fei-Fei, ``Image{N}et: A
  large-scale hierarchical image database,'' in \emph{Proc. CVPR}, 2009.

\bibitem{xiao2010}
J.~Xiao, J.~Hays, K.~A. Ehinger, A.~Oliva, and A.~Torralba, ``{SUN} database:
  Large-scale scene recognition from abbey to zoo,'' in \emph{Proc. CVPR},
  2010.

\bibitem{lin2014}
T.-Y. Lin, M.~Maire, S.~Belongie, J.~Hays, P.~Perona, D.~Ramanan,
  P.~Doll{\'a}r, and C.~L. Zitnick, ``Microsoft {COCO}: Common objects in
  context,'' in \emph{Proc. ECCV}, 2014.

\bibitem{krishna2017}
R.~Krishna, Y.~Zhu, O.~Groth, J.~Johnson, K.~Hata, J.~Kravitz, S.~Chen,
  Y.~Kalantidis, L.-J. Li, D.~A. Shamma \emph{et~al.}, ``Visual genome:
  Connecting language and vision using crowdsourced dense image annotations,''
  \emph{IJCV}, 2017.

\bibitem{kuznetsova2020}
A.~Kuznetsova, H.~Rom, N.~Alldrin, J.~Uijlings, I.~Krasin, J.~Pont-Tuset,
  S.~Kamali, S.~Popov, M.~Malloci, A.~Kolesnikov \emph{et~al.}, ``The {O}pen
  {I}mages dataset {V}4,'' \emph{IJCV}, 2020.

\bibitem{he2016}
K.~He, X.~Zhang, S.~Ren, and J.~Sun, ``Deep residual learning for image
  recognition,'' in \emph{Proc. CVPR}, 2016.

\bibitem{tan2019}
M.~Tan and Q.~Le, ``Efficient{N}et: Rethinking model scaling for convolutional
  neural networks,'' in \emph{Proc. ICML}, 2019.

\bibitem{dosovitskiy2021}
A.~Dosovitskiy, L.~Beyer, A.~Kolesnikov, D.~Weissenborn, X.~Zhai,
  T.~Unterthiner, M.~Dehghani, M.~Minderer, G.~Heigold, S.~Gelly \emph{et~al.},
  ``An image is worth 16x16 words: Transformers for image recognition at
  scale,'' \emph{arXiv preprint arXiv:2010.11929}, 2021.

\bibitem{tolstikhin2021}
I.~Tolstikhin, N.~Houlsby, A.~Kolesnikov, L.~Beyer, X.~Zhai, T.~Unterthiner,
  J.~Yung, D.~Keysers, J.~Uszkoreit, M.~Lucic \emph{et~al.}, ``{MLP}-{M}ixer:
  An all-{MLP} architecture for vision,'' in \emph{Proc. NeurIPS}, 2021.

\bibitem{ferrand2020}
{\'{E}}.~L. Ferrand, S.~Bird, and L.~Besacier, ``Enabling interactive
  transcription in an indigenous community,'' in \emph{Proc. Coling}, 2020.

\bibitem{oquab2015}
M.~Oquab, L.~Bottou, I.~Laptev, and J.~Sivic, ``Is object localization for
  free? weakly-supervised learning with convolutional neural networks,'' in
  \emph{Proc. CVPR}, 2015.

\bibitem{selvaraju2017}
R.~Selvaraju, M.~Cogswell, A.~Das, R.~Vedantam, D.~Parikh, and D.~Batra,
  ``Grad-{CAM}: {Visual} {Explanations} from {Deep} {Networks} via
  {Gradient}-{Based} {Localization},'' in \emph{Proc. ICCV}, 2017.

\bibitem{peddinti2015}
V.~Peddinti, D.~Povey, and S.~Khudanpur, ``A time delay neural network
  architecture for efficient modeling of long temporal contexts,'' in
  \emph{Proc. Interspeech}, 2015.

\bibitem{dai2017}
W.~Dai, C.~Dai, S.~Qu, J.~Li, and S.~Das, ``Very deep convolutional neural
  networks for raw waveforms,'' in \emph{Proc. ICASSP}, 2017.

\bibitem{shon2018}
S.~Shon, H.~Tang, and J.~Glass, ``Frame-level speaker embeddings for
  text-independent speaker recognition and analysis of end-to-end model,'' in
  \emph{Proc. SLT}, 2018.

\bibitem{snyder2018}
D.~Snyder, D.~Garcia-Romero, G.~Sell, D.~Povey, and S.~Khudanpur, ``X-vectors:
  Robust {DNN} embeddings for speaker recognition,'' in \emph{Proc. ICASSP},
  2018.

\bibitem{olaleye2020}
K.~Olaleye, B.~van Niekerk, and H.~Kamper, ``Towards localisation of keywords
  in speech using weak supervision,'' \emph{SAS NeurIPS Workshop}, 2020.

\bibitem{olaleye2021}
K.~Olaleye and H.~Kamper, ``Attention-based keyword localisation in speech
  using visual grounding,'' in \emph{Proc. Interspeech}, 2021.

\bibitem{chrupala2021}
G.~Chrupa{\l}a, ``Visually grounded models of spoken language: A survey of
  datasets, architectures and evaluation techniques,'' \emph{arXiv preprint
  arXiv:2104.13225}, 2021.

\bibitem{peng2021}
P.~Peng and D.~Harwath, ``Fast-slow transformer for visually grounding
  speech,'' \emph{arXiv preprint arXiv:2109.08186}, 2021.

\bibitem{drexler2017}
J.~Drexler and J.~Glass, ``Analysis of audio-visual features for unsupervised
  speech recognition,'' in \emph{Proc. GLU}, 2017.

\bibitem{scholten2021}
S.~Scholten, D.~Merkx, and O.~Scharenborg, ``Learning to recognise words using
  visually grounded speech,'' in \emph{Proc. ISCAS}, 2021.

\bibitem{kamper2018}
H.~Kamper and M.~Roth, ``Visually grounded cross-lingual keyword spotting in
  speech,'' in \emph{Proc. SLTU}, 2018.

\bibitem{wang2021b}
X.~Wang, J.~van~der Hout, J.~Zhu, M.~Hasegawa-Johnson, and O.~Scharenborg,
  ``Synthesizing spoken descriptions of images,'' \emph{IEEE/ACM TASLP}, 2021.

\bibitem{socher2010}
R.~Socher and L.~Fei-Fei, ``Connecting modalities: Semi-supervised segmentation
  and annotation of images using unaligned text corpora,'' in \emph{Proc.
  CVPR}, 2010.

\bibitem{weston2011}
J.~Weston, S.~Bengio, and N.~Usunier, ``Wsabie: Scaling up to large vocabulary
  image annotation,'' in \emph{IJCAI}, 2011.

\bibitem{farhadi2010}
A.~Farhadi, M.~Hejrati, M.~A. Sadeghi, P.~Young, C.~Rashtchian, J.~Hockenmaier,
  and D.~Forsyth, ``Every picture tells a story: Generating sentences from
  images,'' in \emph{Proc. ECCV}, 2010.

\bibitem{kulkarni2011}
G.~Kulkarni, V.~Premraj, S.~Dhar, S.~Li, Y.~Choi, A.~C. Berg, and T.~L. Berg,
  ``Baby talk: Understanding and generating simple image descriptions,'' in
  \emph{Proc. CVPR}, 2011.

\bibitem{young2014}
P.~Young, A.~Lai, M.~Hodosh, and J.~Hockenmaier, ``From image descriptions to
  visual denotations: New similarity metrics for semantic inference over event
  descriptions,'' \emph{TACL}, 2014.

\bibitem{karpathy2015}
A.~Karpathy and L.~Fei-Fei, ``Deep visual-semantic alignments for generating
  image descriptions,'' in \emph{Proc. CVPR}, 2015.

\bibitem{bernardi2016}
R.~Bernardi, R.~Cakici, D.~Elliott, A.~Erdem, E.~Erdem, N.~Ikizler-Cinbis,
  F.~Keller, A.~Muscat, and B.~Plank, ``Automatic description generation from
  images: A survey of models, datasets, and evaluation measures,'' \emph{JAIR},
  2016.

\bibitem{sharma2018}
P.~Sharma, N.~Ding, S.~Goodman, and R.~Soricut, ``Conceptual captions: A
  cleaned, hypernymed, image alt-text dataset for automatic image captioning,''
  in \emph{Proc. ACL}, 2018.

\bibitem{liu2017}
C.~Liu, J.~Mao, F.~Sha, and A.~Yuille, ``Attention correctness in neural image
  captioning,'' in \emph{Proc. AAAI}, 2017.

\bibitem{lu2018}
J.~Lu, J.~Yang, D.~Batra, and D.~Parikh, ``Neural baby talk,'' in \emph{Proc.
  CVPR}, 2018.

\bibitem{Rohrbach2018}
A.~Rohrbach, L.~A. Hendricks, K.~Burns, T.~Darrell, and K.~Saenko, ``Object
  hallucination in image captioning,'' in \emph{Proc. EMNLP}, 2018.

\bibitem{selvaraju2019}
R.~R. Selvaraju, S.~Lee, Y.~Shen, H.~Jin, D.~Batra, and D.~Parikh, ``Taking a
  hint: Leveraging explanations to make vision and language models more
  grounded,'' in \emph{Proc. ICCV}, 2019.

\bibitem{plummer2015}
B.~A. Plummer, L.~Wang, C.~M. Cervantes, J.~C. Caicedo, J.~Hockenmaier, and
  S.~Lazebnik, ``Flickr30k entities: Collecting region-to-phrase
  correspondences for richer image-to-sentence models,'' \emph{IJCV}, 2015.

\bibitem{ponttuset2020}
J.~Pont-Tuset, J.~Uijlings, S.~Changpinyo, R.~Soricut, and V.~Ferrari,
  ``Connecting vision and language with localized narratives,'' in \emph{Proc.
  ECCV}, 2020.

\bibitem{hinton2015}
G.~Hinton, O.~Vinyals, and J.~Dean, ``Distilling the knowledge in a neural
  network,'' \emph{arXiv preprint arXiv:1503.02531}, 2015.

\bibitem{ba2013}
L.~J. Ba and R.~Caruana, ``Do deep nets really need to be deep?'' in
  \emph{Proc. NeurIPS}, 2013.

\bibitem{gou2021}
J.~Gou, B.~Yu, S.~J. Maybank, and D.~Tao, ``Knowledge distillation: A survey,''
  \emph{IJCV}, 2021.

\bibitem{gupta2016}
S.~Gupta, J.~Hoffman, and J.~Malik, ``Cross modal distillation for supervision
  transfer,'' in \emph{Proc. CVPR}, 2016.

\bibitem{zhao2018}
M.~Zhao, T.~Li, M.~Abu~Alsheikh, Y.~Tian, H.~Zhao, A.~Torralba, and D.~Katabi,
  ``Through-wall human pose estimation using radio signals,'' in \emph{Proc.
  CVPR}, 2018.

\bibitem{aytar2016}
Y.~Aytar, C.~Vondrick, and A.~Torralba, ``Sound{N}et: Learning sound
  representations from unlabeled video,'' in \emph{Proc. NeurIPS}, 2016.

\bibitem{albanie2018}
S.~Albanie, A.~Nagrani, A.~Vedaldi, and A.~Zisserman, ``Emotion recognition in
  speech using cross-modal transfer in the wild,'' in \emph{Proc. ACM
  Multimedia}, 2018.

\bibitem{owens2016phd}
A.~H. Owens, ``Learning visual models from paired audio-visual examples,''
  Ph.D. dissertation, 2016.

\bibitem{owens2018}
A.~Owens, J.~Wu, J.~H. McDermott, W.~T. Freeman, and A.~Torralba, ``Learning
  sight from sound: Ambient sound provides supervision for visual learning,''
  \emph{IJCV}, 2018.

\bibitem{alwassel2020}
H.~Alwassel, D.~Mahajan, B.~Korbar, L.~Torresani, B.~Ghanem, and D.~Tran,
  ``Self-supervised learning by cross-modal audio-video clustering,'' in
  \emph{Proc. NeurIPS}, 2020.

\bibitem{viola2001}
P.~Viola, M.~Jones \emph{et~al.}, ``Robust real-time object detection,''
  \emph{IJCV}, 2001.

\bibitem{zeiler2014}
M.~D. Zeiler and R.~Fergus, ``Visualizing and understanding convolutional
  networks,'' in \emph{Proc. ECCV}, 2014.

\bibitem{kamper2019a}
H.~Kamper, A.~Anastassiou, and K.~Livescu, ``Semantic query-by-example speech
  search using visual grounding,'' in \emph{Proc. ICASSP}, 2019.

\bibitem{zhou2016}
B.~Zhou, A.~Khosla, A.~Lapedriza, A.~Oliva, and A.~Torralba, ``Learning deep
  features for discriminative localization,'' in \emph{Proc. CVPR}, 2016.

\bibitem{graves2014}
A.~Graves, G.~Wayne, and I.~Danihelka, ``Neural turing machines,'' \emph{arXiv
  preprint arXiv:1410.5401}, 2014.

\bibitem{luong2015}
M.-T. Luong, H.~Pham, and C.~D. Manning, ``Effective approaches to
  attention-based neural machine translation,'' in \emph{Proc. EMNLP}, 2015.

\bibitem{tamer2020}
N.~C. Tamer and M.~Sara{\c{c}}lar, ``Keyword search for sign language,'' in
  \emph{Proc. ICASSP}, 2020.

\bibitem{ilse2018}
M.~Ilse, J.~Tomczak, and M.~Welling, ``Attention-based deep multiple instance
  learning,'' in \emph{Proc. ICML}, 2018.

\bibitem{simonyan2014}
K.~Simonyan and A.~Zisserman, ``Very deep convolutional networks for
  large-scale image recognition,'' \emph{arXiv preprint arXiv:1409.1556}, 2014.

\bibitem{park2019}
D.~S. Park, W.~Chan, Y.~Zhang, C.-C. Chiu, B.~Zoph, E.~D. Cubuk, and Q.~V. Le,
  ``Specaugment: A simple data augmentation method for automatic speech
  recognition,'' in \emph{Proc. Interspeech}, 2019.

\bibitem{kingma2015}
D.~Kingma and J.~Ba, ``Adam: A method for stochastic optimization,'' in
  \emph{Proc. ICLR}, 2015.

\bibitem{van2022}
E.~van~der Westhuizen, H.~Kamper, R.~Menon, J.~Quinn, and T.~Niesler, ``Feature
  learning for efficient {ASR}-free keyword spotting in low-resource
  languages,'' \emph{Comput. Speech Lang.}, 2022.

\end{thebibliography}

\end{document}